\documentclass[journal]{IEEEtran}
\usepackage{graphicx}
\usepackage{amsmath}
\usepackage{amssymb}
\usepackage{subfigure}
\usepackage{stfloats}
\usepackage{bm}
\usepackage{bbm}
\usepackage{algorithm}
\usepackage{algorithmicx}
\usepackage{algpseudocode}
\usepackage{bm}
\usepackage{tabularx}
\usepackage{array}
\usepackage{multirow}
\usepackage{color}
\usepackage{url}
\usepackage[linkcolor=black,citecolor=black,urlcolor=green,colorlinks=true]{hyperref}
\usepackage{verbatim}
\usepackage{setspace}
\usepackage{mathrsfs}
\usepackage{cite}

\begin{document}
\title{CVR-LSE: Compact Vectorization Representation of Local Static Environments for Unmanned Ground Vehicles}
\author{\IEEEauthorblockN{Haiming Gao, Qibo Qiu, Wei Hua, Xuebo Zhang,~\IEEEmembership{Senior Member,~IEEE}, Zhengyong Han, Shun Zhang}
	\thanks{This work is supported in part by the Key Research and Development Program of Zhejiang Province (No. 2021C01012), in part by Tianjin Science Fund for Distinguished Young Scholars (No. 19JCJQJC62100), and in part by Tianjin Natural Science Foundation (No. 19JCYBJC18500) \emph{(Corresponding author: Wei Hua.)}}%
	\thanks{Haiming Gao, Qibo Qiu, Wei Hua, Zhengyong Han and Shun Zhang are with Zhejiang Lab, Hangzhou, P.~R.~China, 311121 (Email: gaohaiming@zhejianglab.com; huawei@zhejianglab.com).}
	\thanks{Xuebo Zhang is with the Institute of Robotics and Automatic Information System (IRAIS), Tianjin Key Laboratory of Intelligent Robotics (TJKLIR), Nankai University, Tianjin, P.~R.~China, 300350.}%
}
	
\maketitle
	
\begin{abstract}
According to the requirement of general static obstacle detection, this paper proposes a compact vectorization representation approach of local static environments for unmanned ground vehicles. At first, by fusing the data of LiDAR and IMU, high-frequency pose information is obtained. Then, through the two-dimensional (2D) obstacle points generation, the process of grid map maintenance with a fixed size is proposed. Finally, the local static environment is described via multiple convex polygons, which is realized throungh the double threshold-based boundary simplification and the convex polygon segmentation. Our proposed approach has been applied in a practical driverless project in the park, and the qualitative experimental results on typical scenes verify the effectiveness and robustness. In addition, the quantitative evaluation shows the superior performance on making use of fewer number of points information (decreased by about $60\%$) to represent the local static environment compared with the traditional grid map-based methods. Furthermore, the performance of running time ($15ms$) shows that the proposed approach can be used for real-time local static environment perception. The corresponding code can be accessed at {\url{https://github.com/ghm0819/cvr_lse}}.
\end{abstract}

\IEEEpeerreviewmaketitle
	
\section{Introduction}

\IEEEPARstart{A}{utonomous} vehicle is a hot field at present, and the related research have made great progress in the past decade\cite{Ghorai}. In many practical applications, with the aid of simultaneous localization and mapping (SLAM)\cite{Bresson} and High Definition (HD) Map construction \cite{Xiong}, the workspace of the vehicles could be obtained as a \emph{priori}. Thus robust local environment perception is one of the most challenging task, including dynamic/static obstacles detection.

Robust obstacle detection based on various exteroceptive sensors is an indispensable function of unmanned ground vehicles in the process of performing various tasks \cite{Cherubini,CQWang,XZhang,JWen}. Exteroceptive sensors, such as camera, LiDAR and RADAR, etc., are used to obtain the appearance and structure information of the environment, and LiDAR is one of the most popular exteroceptive sensors in driverless vehicles for its effectiveness for mid-near range and multi-target object detection \cite{Ghorai}. In particular, with the rise of vehicle assistant driving and the development of new LiDAR technology, the use cost of LiDAR sensors has shown an obvious downward trend.

According to whether the obstacles are static or not, obstacle types are generally classified into two categories: dynamic obstacle and static obstacle \cite{HZhu}. For the former, it mainly includes vehicles, pedestrians and bicycles, etc., and an important progress has been made for dynamic obstacle detection in the past period of time, especially deep learning-based methods \cite{VNguyen,Shepel,Theodose,QWang} have achieved high accurate detection results. On the other hand, for static obstacles in complex environments, several approaches on general obstacle detection \cite{Garnett,Asvadi,Schreier} inspired by the robotic community have been proposed.

Although those aforementioned methods have made a great progress in the field of autonomous vehicles, the research on robust local static environment perception is still far from mature \cite{Ghorai,HZhu}. Especially for all kinds of static obstacles never seen before in complex urban environments, which could not be detected robustly through deep learning-based approaches \cite{KLi}. In addition, several grid map-based approaches \cite{Danescu,Tanzmeister1,Tanzmeister2,Steyer,YMin} have been proposed to overcome the problem of general obstacle detection. However, considering the requirements of future motion planning of vehicles \cite{Ziegler,CLiu}, and traditional occupancy grid map has known disadvantages of consuming a lot of memory and increasing the computational cost \cite{BMu}. On the other hand, for the unmanned ground vehicle whose workspace is on the two-dimensional (2D) plane, it is enough to use the 2D grid map with the fixed size to describe the local environment. Therefore, robust and compact general obstacle detection method in local environment is very indispensable for autonomous vehicles.

Motivated by the requirement of robust and compact local environment perception mentioned above, we propose a novel and compact vectorization representation approach of local static environments for unmanned ground vehicles, called VR-LSE. Firstly, we make use of the LiDAR data and the IMU data to obtain the high-frequency vehicle pose through pose interpolation. Then, the 2D obstacle points are generated through ground plane extraction and nearest obstacle point selection, which is inspired by the idea of grid map generation based on the 2D LiDAR sensor. Finally, combined with the respective advantages of grid map in multi-frame fusion and convex polygons in simplified expression, a new general obstacle detection approach is proposed for local static environments. Our proposed approach is applied in our driverless project in the park, which is good at general obstacle detection, and four typical scenes on general obstacles never seen before show the effectiveness of the proposed approach. In addition, the quantitative evaluation is carried out to show the superior performance in terms of the simplification for static obstacle representation, and the performance of running time shows that the proposed approach can be used for real-time local static environment perception. The main contributions of this paper include the following three aspects:

(1) Through 2D obstacle points generation, we make full use of the advantage of grid map in multi-frame fusion to generate the robust representation of static obstacles around the vehicle. In addition, the grid map is maintained within a fixed range as the vehicle moves on, which is used to reduce memory consumption and reuse the map information.

(2) We propose a compact vectorization representation approach of local static environments, which makes use of the multiple convex polygons to describe the general static obstacles through double threshold-based boundary simplification and convex polygon segmentation. Therefore, the description of local static environment is more effective and lightweight with the less information.

(3) The proposed general obstacle detection approach has been applied in the driverless project in the park, which presents excellent performance at all kinds of static obstacles never seen before and is a sufficient supplement to the deep learning-based methods.

The remaining parts of this paper are structured as follows. Section II describes related works of obstacle detection in the fields of autonomous vehicles. In the next section, we present the proposed the novel and compact approach called VR-LSE in details. Section IV provides the typical cases and quantitative evaluation to show the effectiveness of VR-LSE. Finally, the paper is concluded in Section V.

\section{Related work}
Robust obstacle detection plays a more and more important role in the process of automatic driving, especially for the safety and reliability of autonomous vehicles. In this section, we briefly discuss previous related works in the fields of obstacle detection and the related grid map.

\subsection{Obstacle detection}
According to the different implementations, obstacle detection can be divided into two specific categories: Based on traditional methods and based on deep learning. In \cite{ZZhang}, Zhang et al. make use of the stereo camera system to detect and track dynamic obstacles, and the distance information of all obstacles is obtained through disparity computation and triangulation. Based on the same stereo camera system, the work in \cite{Bovcon} proposes a new obstacle detection algorithm for unmanned surface vehicles, especially a stereo verification scheme to reduce the false positive and false negative detections. In addition, the stereo frame is also used for scene understanding and object discovery in dynamic street scenes \cite{Kochanov}. On the other hand, in recent years, deep learning-based approaches have been widely applied. In \cite{Garnett}, the authors a unified network with real-time detection capability for both categorized and uncategorized object, and the network is trained through both manually and automatically generated annotations using LiDAR. Nguyen et al. \cite{VNguyen} propose a general framework for dynamic obstacle detection and tracking, which is realized based on a novel deep learning approach with the use of multiple sources of local patterns and depth information. By occupancy map generation, the work in \cite{Shepel} presents a new algorithm for detection of dynamic and static obstacles around the vehicle from noisy point clouds obtained from the stereo camera system.

Compared with the camera sensors, LiDAR sensors could provide more precise and reliable structure information of the environments, which are applied widely for obstacle detection of autonomous vehicles. Choi et al. \cite{Choi} present a unified framework for solving two environmental perception tasks concurrently: SLAM and moving-object tracking (MOT), and MOT is realized through the geometric model-based tracking with multiple motion models. In \cite{Bersani}, Bersani et al. propose a novel state estimation approach for autonomous driving, which provides real-time state estimates for the ego-vehicle and the surrounding obstacles. In order to overcome the spatial non-uniformity of LiDAR data, the work in \cite{FGao} proposes a dynamic clustering algorithm by introducing an ellipse neighbor for better clustering performances. In addition, the 2D LiDAR sensor is also used in autonomous vehicles, Dong et al. \cite{HDong} we propose a real-time and reliable approach to detect and track obstacles, which are represented by a set of points against their outlines. And Lee et al. \cite{HHLee} present a novel model-free approach for detection and tracking of moving objects (DATMO), and the real-time application is available via Static Obstacle Map (SOM) and Geometric-Model-Free Approach (GMFA).

On the other hand, learning-based approaches with 3D LiDAR have also been applied widely for autonomous vehicle. The work in \cite{XWu} presents a novel multi-modal framework called SFD (Sparse Fuse Dense), which makes use of pseudo point clouds to tackle the difference between the images and the point clouds, and realize high-quality 3D obstacle detection. In order to overcome the dependency of existing neural network-based approaches on point cloud resolution, Th{\'e}odose et al. propose an approach to improve performances over unknown data without supplementary knowledge, which has important reference significance for uncertainty estimation in the field of automatic vehicles. In \cite{QWang}, a pose estimation network based on the LiDAR sensor is proposed to estimate the poses of both the ego vehicle and the obstacles, in which PointNet++ \cite{CQi} is used to extract point-wise features and divide the points into the static and the dynamic.

Although the above obstacle detection approaches have achieved prominent success in the field of autonomous vehicles, they are generally incapable of detecting several novel objects that are not seen before.

\subsection{Grid map}
Because of the advantages of multi-frame fusion and easy construction, grid map has been utilized widely in robotic community \cite{Thrun}. and in recent years, the grid map has been applied in the field of obstacle detection and tracking for autonomous vehicles, especially for unknown general obstacles. In \cite{CWang}, combined with SLAM and moving object tracking, a mathematical framework is established, and two solutions are described: SLAM with generalized objects, and SLAM with detection and tracking of moving objects (DATMO). In order to obtain the accurate grid map of the static environment with fast computation, Homm et al. \cite{Homm} make use of the Graphics Processing Unit (GPU) to overcome the limitations of classical occupancy grid computation. In addition, grid map has also been used to detect and track dynamic obstacles, in \cite{Danescu}, the authors present a novel occupancy grid tracking solution based on particles for tracking the dynamic driving environment. Tanzmeister et al. \cite{Tanzmeister1,Tanzmeister2} propose grid-based tracking and mapping (GTAM) approach, which is based on the low-level grid and simultaneously estimates the static and the dynamic environments. Based on the the Dempster-Shafer framework, the work in \cite{Steyer} proposes a new dynamic grid mapping approach, which is used to model hypotheses for static occupancy, dynamic occupancy, free space, and their combined hypotheses. In addition to the 2D grid map, Min et al. \cite{YMin} propose a 3-D dynamic occupancy mapping approach, called K3DOM, which is realized through adapting kernel inference for dynamic environments with particle tracking. On the other hand, through combining with the stereo camera system and the deep neural network FCN-ResNet-M-OC, the authors in \cite{Shepel} propose a novel approach for generating an occupancy grid from a semantic point cloud.

However traditional grid map-based approach have known disadvantages of consuming a lot of memory and increasing the computational cost of future motion planning.

\section{Local Static Environment Representation}

\begin{figure}[!htb]
	\centering
	\includegraphics[width=8cm]{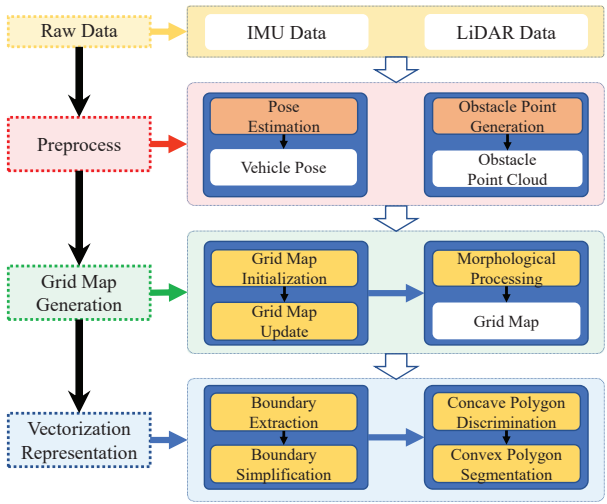}
	\caption{Schematic illustration of the the process of vectorization representation approach for local static environments. The yellow module represent sensor information, including LiDAR data and IMU data. In addition, the red module, blue module and green module represent the main processes, respectively, which are detailed in Section A-D.}
	\label{overallframework}
\end{figure}

This section introduces the framework and the detailed procedures of the proposed static environmental representation approach for unmanned ground vehicles, as shown in Fig. \ref{overallframework}. The input sensor information includes the IMU data and the LiDAR data, which is marked by the yellow rectangle. On this basis, the vehicle pose and the obstacle point cloud are obtained via the {\textbf{Pose Estimation}} module and the {\textbf{Ground Segmentation}} module, respectively. Then, the {\textbf{Grid Map Generation}} module is carried to obtain the final grid map information for local environments based on the vehicle pose and the obstacle point cloud. Finally, several convex polygons are used to represent the local static environment through the {\textbf{Vectorization Representation}} module.

\subsection{Pose estimation}

\begin{figure}[!htb]
	\centering
	\includegraphics[width=8cm]{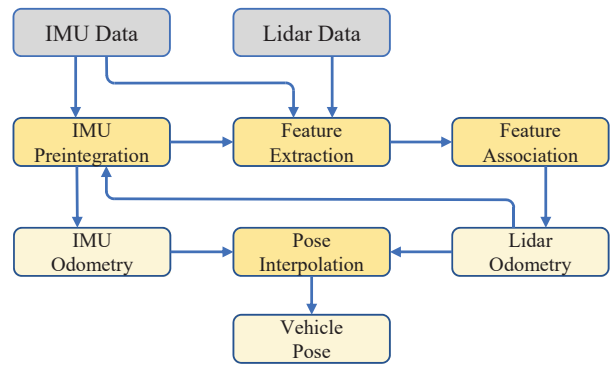}
	\caption{Schematic illustration of the process of pose estimation process using LiDAR and IMU. The gray modules represent the sensor data, and the yellow modules are main processes of pose estimation.}
	\label{pose_estimation}
\end{figure}

In the sensor data preprocess stage, the real-time vehicle pose is obtained. Inspired by the well-known LiDAR SLAM solution called LIO-SAM\cite{TixiaoShan}, the pose estimation method for unmanned ground vehicles is proposed based on IMU data and LiDAR data in this section. Considering the vehicle pose is only used to update and obtain the local static environments, different from the classical LiDAR SLAM process, the loop closure detection and global map maintenance are not needed the process of pose estimation.

Therefore, the whole process of pose estimation is shown as Fig. \ref{pose_estimation}. Firstly, based on the IMU data and the LiDAR odometry, the process of IMU Preintegration is carried out, which is used the estimate the bias of the IMU sensor and provide the IMU odometry. Then, based on the IMU data and point cloud, edge and planar features are extracted via calculating the roughness for each point combined with neighbourhood points, and the more detailed description of the feature extraction can be found in \cite{TixiaoShan}. In addition, the feature association between the local map and current LiDAR scan is carried out. Note that the local map is obtained through dynamic maintenance, which is consisted of latest $n$ LiDAR key-scans ($n$ is the given threshold, which is set as $50$ in this paper). Finally, combined with IMU odometry and LiDAR odometry, the high-frequency vehicle pose is obtained through pose interpolation.

\subsection{Obstacle point generation}

\begin{figure}[!htb]
	\centering
	\includegraphics[width=8cm]{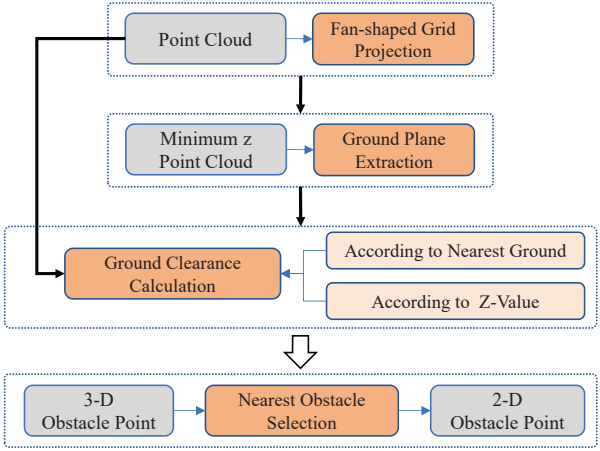}
	\caption{Schematic illustration of the process of obstacle point generation. The gray modules represent data information, and the orange modules are the four main processes of obstacle point extraction.}
	\label{obstacle_point_generation}
\end{figure}

\begin{figure}[!htb]
	\centering
	\includegraphics[width=8.5cm]{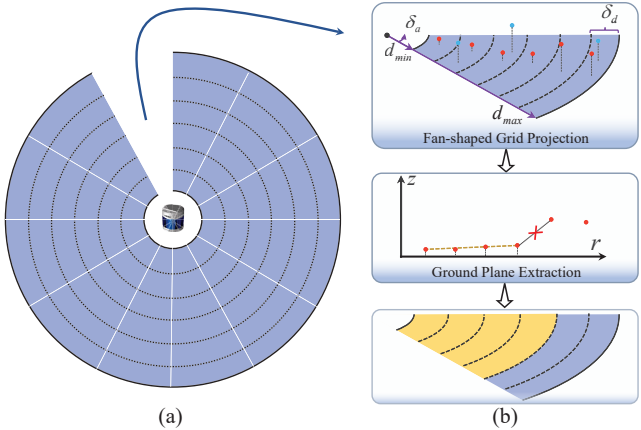}
	\caption{Schematic illustration of fan-shaped grid map (a) and the ground plane extraction process (b). The fan-shaped grid map is consisted of several segments, and each segment is segmented into several bins.}
	\label{fan_shape}
\end{figure}

In the meantime, the obstacle points are generated, which are used to obtain the local static environments combined with the real-time vehicle pose. While the detail process of obstacle point generation is shown as Fig. \ref{obstacle_point_generation}. At first, the whole point cloud is projected into the fan-shaped grid map, as shown in Fig. \ref{fan_shape}(a). The fan-shaped grid map is consisted of $m$ segments, while each segment is segmented into $n$ bins, and the corresponding parameters $m$ and $n$ are calculated as follows,
\begin{align}
	m = {360^\circ} / {\delta_a}, \ \
	n = ({d_{max} - d_{min}}) / {\delta_d},
\end{align}
where $\delta_a$ and $\delta_d$ denote the angle resolution and distance resolution, respectively, while the parameters $d_{max}$ and $d_{min}$ represent the valid perception range. For each point $p_i = (x_i, y_i, z_i)$, the range information $r_i$ and the horizontal angle $\theta_i$ are calculated as follows,
\begin{align}
	r_i = \sqrt{{x_i}^2 + {y_i}^2}, \ \
	\theta_i = atan2(y_i, x_i).
\end{align}

According to the values of $r_i$ and $\theta_i$, the corresponding point $p_i$ is assigned into corresponding $j$-th bin of the $k$-th segment, and the index $k$ and $j$ are calculated as
\begin{align}
	k = [\theta_i / \delta_a], \
	j = [(r_i - d_{min}) / \delta_d] (d_{min} \leqslant r_i \leqslant d_{max}),
\end{align}
while the operator $[ \cdot ]$ represents rounding down. And there are bins without points and bins with multiple points. For the latter, only the point $p_i$ with the minimum value of $z_i$ (marked by the red circle in Fig. \ref{fan_shape}(b)) is applied for the ground plane extraction based on seeded region growing for each segment(Fig. \ref{fan_shape}(b)). Note that the premise of the above approach is that the current position of the unmanned ground vehicle is on the ground plane, while the more detailed information could refer the related work \cite{Himmelsbach}.

More importantly, after obtaining the ground plane representation of each segment (which is denoted by the line segment between the value $z$ and the value $r$, as shown in Fig. \ref{fan_shape}(b)), the ground clearance of each scan point could be calculated. The scan point is assigned into the special bin, and if the corresponding ground plane representation $\left(z = k \cdot r + b\right)$ $(k, b$ are line parameters$)$ exists, then the ground clearance is calculated according to the distance to the corresponding ground plane. Otherwise, the ground clearance is denoted by the $z$-value of the scan point. Specially, the ground clearance $g_i$ of point $p_i$ is calculated as follows,
\begin{align}
	{g_i} = \left\{ {\begin{array}{*{20}{c}}
			{{{\left| {k \cdot {r_i} + b} \right|}}/{{\sqrt {{k^2} + 1} }} \;\;\;\; \text{ground plane exists}}  \\
			{{z_i}\;\;\;\;\;\;\;\;\;\;\;\;\;\;\;\;\; \text{otherwise}}  \\
	\end{array} } \right..
\end{align}

After the above processing of point cloud, the three-dimensional(3D) obstacle points could be obtained. Specifically, the scan point $p_i$ could be represented by $\left(x_i, y_i, g_i\right)$, then it is classified into \emph{ground point} and \emph{obstacle point} according to the ground clearance, and pass-through filters are applied in the ground clearance coordinate to remove ground points and high-altitude obstacle points. The limitation of the pass-through filter must be set according to the size of vehicle.

In addition, considering the workspace of unmanned ground vehicle is in 2D ground plane and inspired by the idea of grid map generation based on the 2D LiDAR sensor, the 3D obstacle points are transform into 2D obstacle points via nearest obstacle selection. Let $\delta_s$ denote the horizontal angular resolution of LiDAR sensor, thus the size of 2D obstacle points is set as $[360^{\circ} / {\delta_s}]$. For each horizontal angle $\alpha_i \ (i \times \delta_s)$ that corresponds the special ray includes multiple obstacle points $(p_1, p_2, ..., p_l)$ with the same horizontal angle, and the final 2D obstacle point ${\cal P}_i = \left(x_i, y_i\right)$ is represented by the obstacle point with the minimum range distance.

\subsection{Grid map generation}
Accordingly, the raw 3D point cloud is converted into a 2D obstacle point cloud ${\cal P} = ({\cal P}_1, {\cal P}_2,..., {\cal P}_s)$, while the size $s$ is set as $[360^{\circ} / {\delta_s}]$. Note that if there is no scan point in the $i$-th ray $R_i$, then the corresponding obstacle point ${\cal P}_i$ is set as $\left({d_{max}}\cos(\alpha_i), {d_{max}}\sin(\alpha_i)\right)$, which is called virtual obstacle point in this section.

\begin{figure}[!htb]
	\centering
	\includegraphics[width=6.5cm]{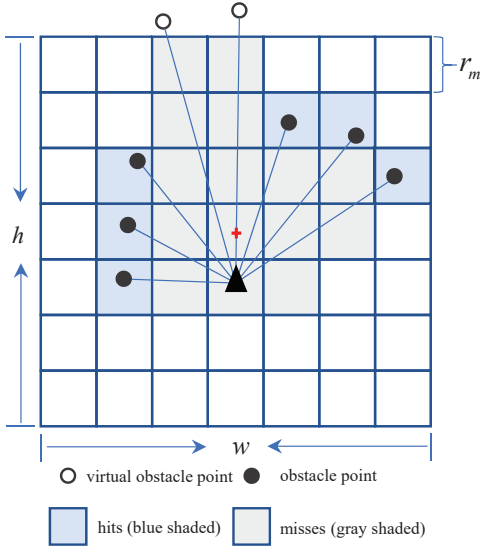}
	\caption{Schematic illustration of obstacle points and grids associated with hits and misses.}
	\label{grid_map}
\end{figure}

Based on the obtained 2D obstacle points, the local static environment is represented through the occupancy grid map, and the process of grid map generation is described below. At first, the grid map initialization is performed, which is built based on the given resolution $r_m$ and size $s_m = \left(w, h\right)$ of the grid map, as shown in Fig. \ref{grid_map}. The map center is marked by the red cross, which does not coincide with the vehicle center (marked by the gray triangle), because forward environmental perception is more important than backward. More importantly, the vehicle center does not coincide with the sensor center in practical application, thus extrinsic parameters need to be calibrated. For simplicity, the sensor center is considered coincidence with the vehicle center in this section.

Certain problems in field of unmanned vehicles are best formulated as estimation problems with binary state (free or occupancy) that does not change over time, especially the problem of local static environment perception involved in this paper. Accordingly, in order to represent the static environment around the vehicle, the binary Bayes filter is applied in grid map generation. The environment is represented as the grid map, thus the environment state estimation problem is converted into the problem of every cell state estimation. Thus the state estimation problem is denoted as
\begin{align}
	p\left( {{\cal M}|{z_{0:t}},{x_{0:t}}} \right) = \prod\limits_{i = 1}^N {p\left( {{c_i}|{z_{0:t}},{x_{0:t}}} \right)},
\end{align}
where $\cal M$ represents the grid map, and the $i$-th cell is denoted by $c_i$, while ${z_{0:t}}$ and ${x_{0:t}}$ are the LiDAR observation information and the vehicle pose information, respectively. In Fig. \ref{grid_map}, whenever the obtained obstacle point set $\cal P$ is to be inserted into the grid map, the cells for hits and for misses are calculated through connecting the obstacle point and the sensor origin. For the $i$-th cell $c_i$ in $\cal M$, the occupancy probability is denoted by $p\left(c_i\right)$, and free probability is presented by $1 - p\left(c_i\right)$. In addition, the log odds representation method\cite{Thrun} is used to update the state of each cell, and the odds of the state $c_i$ is defined as the ratio of the probability of occupancy divided by the probability of free,
\begin{align}
	odd\left(c_i\right)=\frac{p\left(c_i\right)}{1 - p\left(c_i\right)},
\end{align}
and the log value of above equation is denoted as
\begin{align}
	l\left(c_i\right)=\log{\frac{p\left(c_i\right)}{1 - p\left(c_i\right)}},
\end{align}
thus the occupancy probability could be obtained as follows,
\begin{align}
	p\left(c_i\right)=1 - {\frac{1}{1 + \exp{(l\left(c_i\right))}}}.
\end{align}

In the initialization phase, the occupancy probability $p\left(c_i\right)$ is $0.5$, thus the $l_0\left(c_i\right)$ is $0$ through equation (6-7). Then in the updating phase for time step $t+1$, the log odds representation of $c_i$ is updated as
\begin{align}
	l_{t+1}\left(c_i\right)=l_{t}\left(c_i\right) + \log{\frac{p\left(c_i|z_{t+1}, x_{t+1}\right)}{1 - p\left(c_i|z_{t+1}, x_{t+1}\right)}} - l_0\left(c_i\right),
\end{align}
where $p\left(c_i|z_{t+1}, x_{t+1}\right)$ denotes the occupancy probability based on current observation $z_{t+1}$ and $x_{t+1}$. And the occupancy probability $p\left(c_i|z_{t+1}, x_{t+1}\right)$ is set as $\alpha_{hit}$ and $\alpha_{miss}$ for hits and misses, respectively, where $\alpha_{hit}$ and $\alpha_{miss}$ are constant parameters given in advance. Through the log odds representation, the effective value range of cell $c_i$ can range from ${\rm{ - }}\infty$ to ${\rm{ + }}\infty$ and avoid truncation problems that arise for probabilities close to $0$ or $1$.

\textbf{\emph{Remark 1:}} In order to reduce the adverse effects of low dynamic obstacles on static environment description, we set the upper bound $l_{up}$ and lower bound $l_{low}$ for $l\left(c_i\right)$ in practical application, respectively.

\begin{figure}[!htb]
	\centering
	\includegraphics[width=7cm]{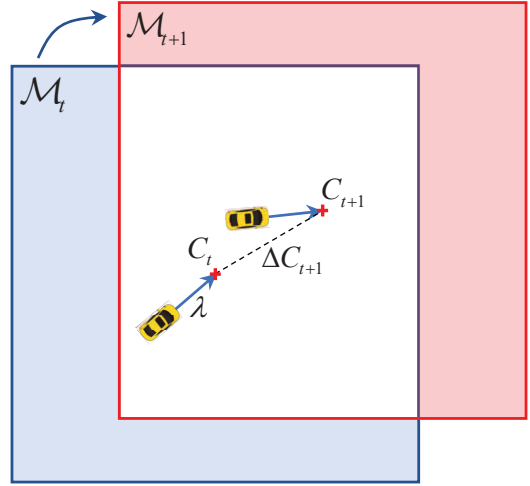}
	\caption{The relationship between the grid map ${\cal M}_{t}$ and the grid map ${\cal M}_{t+1}$.}
	\label{grid_map_move}
\end{figure}

For unmanned ground vehicles operating in large-scale outdoor environments, using the global grid map is an extremely luxurious method due to the limited memory and the irrelevance of old locations. Therefore, we only maintain the local static environment information within the fixed range as the vehicle moves on, which can avoid the loss of map information. At time step $t$, the vehicle pose is denoted as $X_t = (x^v_{t}, y^v_{t}, {\theta}^v_{t})$, and the map center is $C_t = (x^m_{t}, y^m_{t})$, which could be computed as
\begin{align}
	\left\{ {\begin{array}{*{20}{c}}
			{x_t^m = x_t^v + \lambda  \cdot \cos \left( {\theta _t^v} \right)}\\
			{y_t^m = y_t^v + \lambda  \cdot \sin \left( {\theta _t^v} \right)}
	\end{array}} \right. ,
\end{align}
where $\lambda$ denotes the forward distance and the heading angle of grid map is decided by initial vehicle pose, which does not change in the subsequent operation, as shown in Fig. \ref{grid_map_move}. Accordingly, when the vehicle moves at the new pose called $X_{t+1}$ at time step $t+1$, the map center $C_{t+1}$ could be calculated through equation (10); Then move the grid map ${\cal M}_{t}$ of the previous time step to the position of new map center $C_{t+1}$, and the overlap area (marked by the white rectangle in Fig. \ref{grid_map_move}) represent the retained map information, which could be used at time step $t+1$.

\subsection{Vectorization representation}
\subsubsection{Boundary extraction and simplification}
Based on the real-time grid map around the vehicle, this section describes the process of vectorization representation of the static environment in details. At first, in order to ensure the smoothness of environment representation, the submap ${\cal M}_s$ of the overall grid map $\cal M$ is obtained. At the same time, according to equation (8), the occupancy probabilities of all cells in ${\cal M}_s$ could be obtained. Then, we set occupancy threshold ${\beta}_{occ}$ and free threshold ${\beta}_{free}$, and the state of cell $c_i$ could be classified into three categories
as follows,
\begin{align}
	S\left( {{c_i}} \right) = \left\{ {\begin{array}{*{20}{c}}
			{Occupancy}&{p\left( {{c_i}} \right) \ge {\beta _{occ}}}\\
			{Unknown}&{{\beta _{occ}} > p\left( {{c_i}} \right) \ge {\beta _{free}}}\\
			{Free}&{{\beta _{free}} > p\left( {{c_i}} \right)}
	\end{array}} \right..
\end{align}
Afterwards, according to all occupancy cells in ${\cal M}_s$, the binary image $I_s$ could be obtained. Then, morphology processing was performed on the binary image obtained from the previous step to get the segmentation result. On this basis, without making use of clustering methods to obtain the obstacle information, while the traditional method \cite{Suzuki} for boundary detection are applied to extraction the boundary. Compared to the visual image information, the grid map is generated through multi-scans with accurate behaviour information, thus obtained outer boundary could well describe the obstacle information.

Then the double threshold-based boundary simplification approach is proposed to reduce the number of boundary points and improve the efficiency of environment description. Inspired by the line segmentation method called Iterative-End-Point-Fit (IEPF), the proposed boundary simplification approach is shown in \textbf{Algorithm 1}. Let ${\cal C} = \{{\cal {C}}_1,...,{\cal {C}}_l\}$ denotes the obtained boundary set, and ${\cal C'} = \{{\cal {C'}}_1,...,{\cal {C'}}_l\}$ represents the boundary information after processing.

\begin{algorithm}[htb]\footnotesize
	\caption{Double threshold-based boundary simplification}
	\begin{algorithmic}[1]
		\Require Initial boundary set ${\cal C} = \{{\cal {C}}_1,...,{\cal {C}}_l\}$, inner distance threshold $\delta_{in}$, outer distance threshold $\delta_{ou}$,
		minimum number of boundary points $k$.
		\Ensure Boundary set after simplification ${\cal C'} = \{{\cal {C'}}_1,...,{\cal {C'}}_l\}$.
		\For{$i = 1 \to l$, ${\cal {C}}_i \in {\cal C}$}
		\State Initialize the number of points ${\emph{N}}_i$ = size(${\cal {C}}_i$) and the point index pair
		\State queue $I_q$, and insert $(1, {\emph{N}}_i)$ into $I_q$.
		\If{${\emph{N}}_i < k$}
		\State ${\cal {C'}}_i \leftarrow {\cal {C}}_i$
		\EndIf
		\While{$I_q$ is not empty}
		\State Let $d_{in}$ and $d_{ou}$ denote the max inner and outer distance, and
		\State the corresponding index are $I_{in}$ and $I_{ou}$, respectively.
		\State Peek queue front by insertion order, then obtain start index $I_s$
		\State and end index $I_e$.
		\State Initialize the final index set $I_{f} \leftarrow \left\{ {I_s, I_e} \right\}$.
		\State Line parameter calculation through $P_{I_s}$ and $P_{I_e}$.
		\For{$j = (I_s + 1) \to (I_e - 1)$, $P_j \in {\cal C}_i$}
		\State Calculate the distance $d_j$ from point $P_j$ to the line.
		\State Calculate the side attribute $A_j$ of the point $P_j$.
		\If{$A_j$ is $True$ and $d_j > d_{ou}$}
		\State $d_{ou} \leftarrow d_j$
		\State $I_{ou} \leftarrow j$
		\ElsIf{$A_j$ is $false$ and $d_j > d_{in}$}
		\State $d_{in} \leftarrow d_j$
		\State $I_{in} \leftarrow j$
		\EndIf
		\EndFor
		\If{$d_{in} > \delta_{in}$}
		\State Insert $(I_s, I_{in})$ and $(I_{in}, I_e)$ into $I_q$
		\State Insert $I_{in}$ into $I_{f}$
		\ElsIf{$d_{ou} > \delta_{ou}$}
		\State Insert $(I_s, I_{ou})$ and $(I_{ou}, I_e)$ into $I_q$
		\State Insert $I_{ou}$ into $I_{f}$
		\EndIf
		\EndWhile
		\State Sort $I_{f}$ in ascending order according to the index value.
		\State Calculate ${\cal {C'}}_i$ through $I_{f}$ and ${\cal {C}}_i$.
		\EndFor
	\end{algorithmic}
\end{algorithm}

\begin{figure}[!htb]
	\centering
	\includegraphics[width=8.5cm]{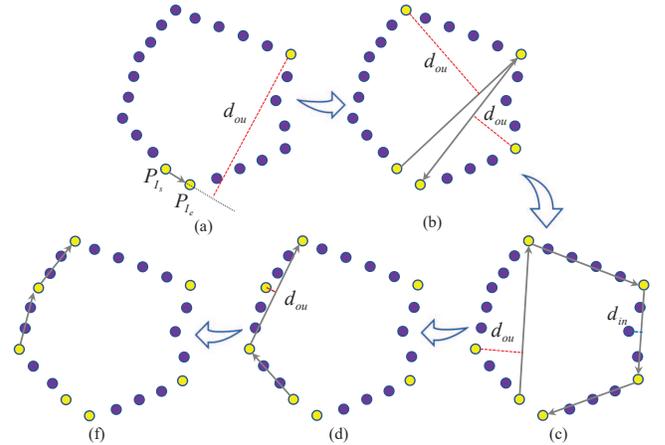}
	\caption{Schematic illustration of double threshold-based boundary simplification. The purple circles are the raw boundary points, and the final boundary points are marked by the yellow circles.}
	\label{line_extraction}
\end{figure}

For the above \textbf{Algorithm 1}, different from the single-threshold-based line fitting methods \cite{Nguyen}, the proposed double-threshold-based simplification approach is used to simplified the extracted boundary, as shown in Fig \ref{line_extraction}. More importantly, compared with the sunken boundary ($d_{in}$ in Fig. \ref{line_extraction}(c)) in the process of obstacle boundary extraction, we pay more attention to the protruding boundary ($d_{ou}$ in Fig. \ref{line_extraction}(d)), which will affect the motion planning of the vehicles. On this basis, the outer distance threshold $\delta_{ou}$ is smaller than the inner distance threshold $\delta_{in}$. Therefore, although $d_{in}$ in Fig. \ref{line_extraction}(c) is bigger than $d_{in}$ in Fig. \ref{line_extraction}(d), the corresponding point is not inserted into the final set $I_f$. Finally, $I_f$ is sorted according to the index value, and the simplified boundary information could be obtained (line 33-34 in \textbf{Algorithm }1).

\subsubsection{Concave polygon discrimination and Convex polygon segmentation}
After the above process of boundary simplification, fewer points can be used for environment description. However, there are convex polygons and concave polygons in the boundary set (${\cal C'} = \{{\cal {C'}}_1,...,{\cal {C'}}_l\}$) of the current description environment, in which concave polygons are not conducive to the rapid development of motion planning. Therefore, the processes of concave polygon discrimination and convex polygon segmentation are carried out, and the final environment description is represented in convex polygon form.

\begin{figure}[!htb]
	\centering
	\includegraphics[width=6cm]{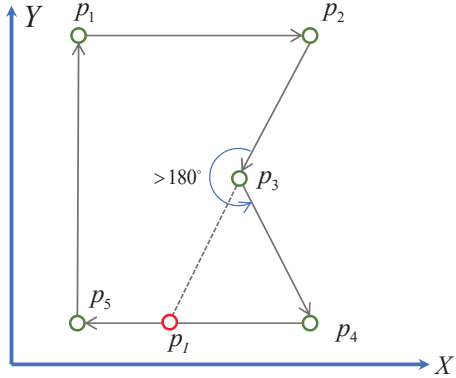}
	\caption{Concave polygon discrimination and Convex polygon segmentation.}
	\label{convex_segmentation}
\end{figure}

As shown in Fig. \ref{convex_segmentation}, we take an example of a boundary consisted of $\{p_1, p_2, ...,p_5\}$ to briefly describe how to discriminate the concave polygon and how to segment the concave polygon into multiple convex polygons. For a polygon, if there is an internal angle whose value is more than 180 degrees, then the polygon is discriminated as a concave polygon. Let $p_{i-1}$, $p_{i}$ and $p_{i+1}$ denote the three consecutive boundary points, in the real application, the current angle can be identified through judging whether $p_{i+1}$ is on the left or right side of the directional line segment composed of $p_{i-1}$ and $p_{i}$, and the corresponding discriminant is expressed as
\begin{align}
	{d_{v}} = \left( {{x_1} - {x_3}} \right) \cdot \left( {{y_2} - {y_3}} \right) - \left( {{y_1} - {y_3}} \right) \cdot \left( {{x_2} - {x_3}} \right),
\end{align}
where if ${d_{v}} > 0$, then the $p_{i+1}$ is on the left side, shown as the $p_4$ in Fig. \ref{line_extraction}, and the corresponding internal angle composed of three points is greater than 180 degrees.

Based on the above, the obtained concave polygon is further segmented into multiple convex polygons. At first, for the concave polygon shown in Fig. \ref{line_extraction}, according to equation (12), the sunken position $\{p_2, p_3, p_4\}$ could be obtained. Then, by extending the edge $(\overrightarrow {{p_2} \ {p_3}})$ until it intersects with another edge ($\overrightarrow {{p_4} \ {p_5}}$), the resulting intersection is denoted as $p_I$, and two polygons $\{p_1, p_2, p_I, p_5\}$ and $\{p_3, p_4, p_I\}$ are obtained after one segmentation. Afterwards, we judge whether there are still sunken positions in the two polygons after segmentation. If not, the process of convex polygon segmentation is finished.

\section{Experiments}
\begin{figure}[!htb]
	\centering
	\includegraphics[width=8.8cm]{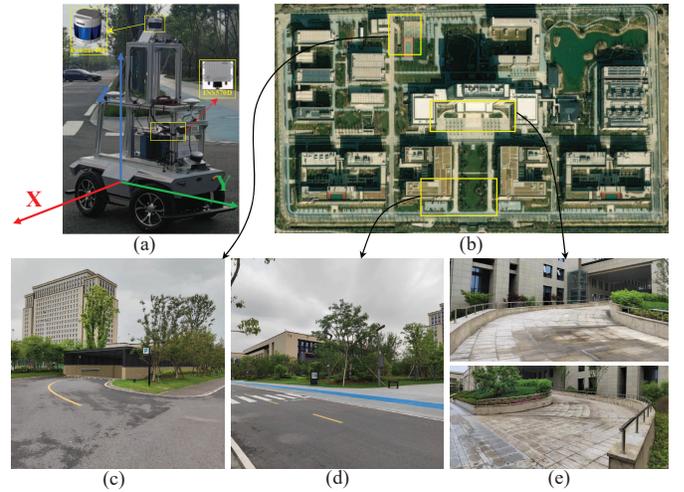}
	\caption{Experimental platform (a) and the experimental environment (b). In addition, (c)-(e) are the typical scenes in the experimental environment.}
	\label{experimental_setup}
\end{figure}

In order to evaluate the performance of the proposed approach, several experiments have been carried out in real environments. The experimental system consists of a ackermann steering ground vehicle and a 3D LiDAR sensor, as shown in Fig \ref{experimental_setup}(a). In addition, the corresponding experimental environment is shown in Fig \ref{experimental_setup}(b), which is a typical park environment including the ground road scene, the underground garage scene and the uphill and downhill scenes. The proposed approach is programmed by C/C++, using a computer with hardware and software specifications listed in Table \ref{hardware}.

\begin{table}[!htb]
	\centering
	\caption{Hardware and software specification}
	\begin{tabular}{c|c}
		\hline \hline
		Vehicle platform           & YUHESEN FR-09  \\
		\hline
		LiDAR Sensor          & Pandar40P \\
		\hline
		Processor       & Intel Core i7-10700K @3.8Ghz \\
		\hline
		RAM             & 16GB  \\
		\hline
		Operating system        & Ubuntu 18.04          \\
		\hline
		ROS             & moletic         \\
		\hline \hline
	\end{tabular}
	\label{hardware}
\end{table}

\begin{figure}[!h]
	\centering
	\includegraphics[width=8.5cm]{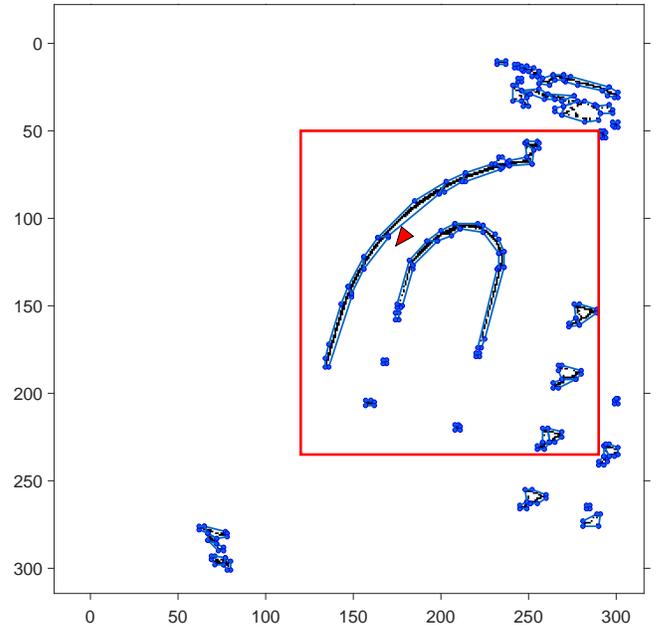}
	\caption{The occupancy grid map and the final convex polygons. The vehicle is marked through the red filled triangle.}
	\label{process_one}
\end{figure}

\begin{figure*}[!h]
	\centering
	\subfigure[Boundary extraction result]{\includegraphics[width=5.8cm]{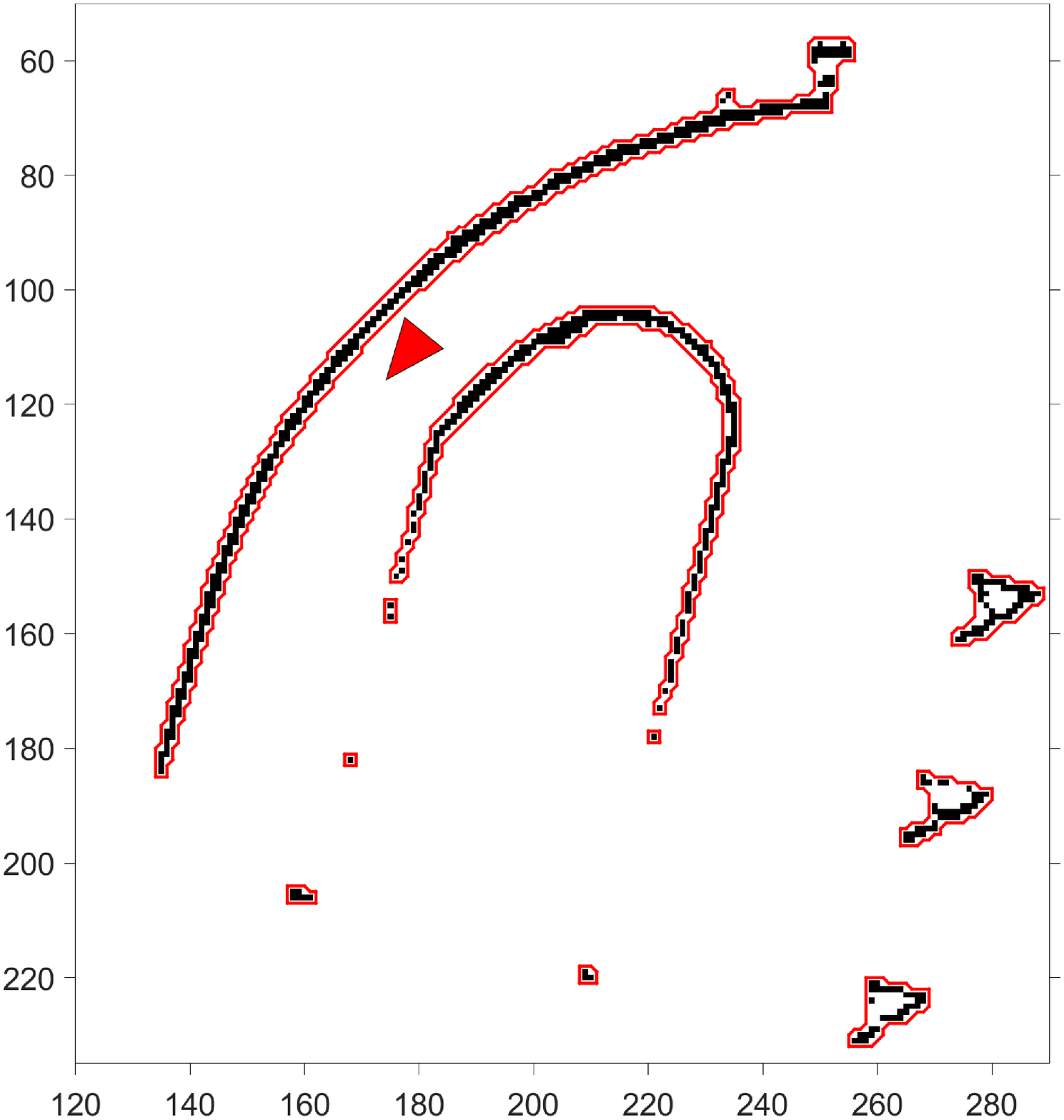}}
	\centering
	\subfigure[Boundary simplification result]{\includegraphics[width=5.8cm]{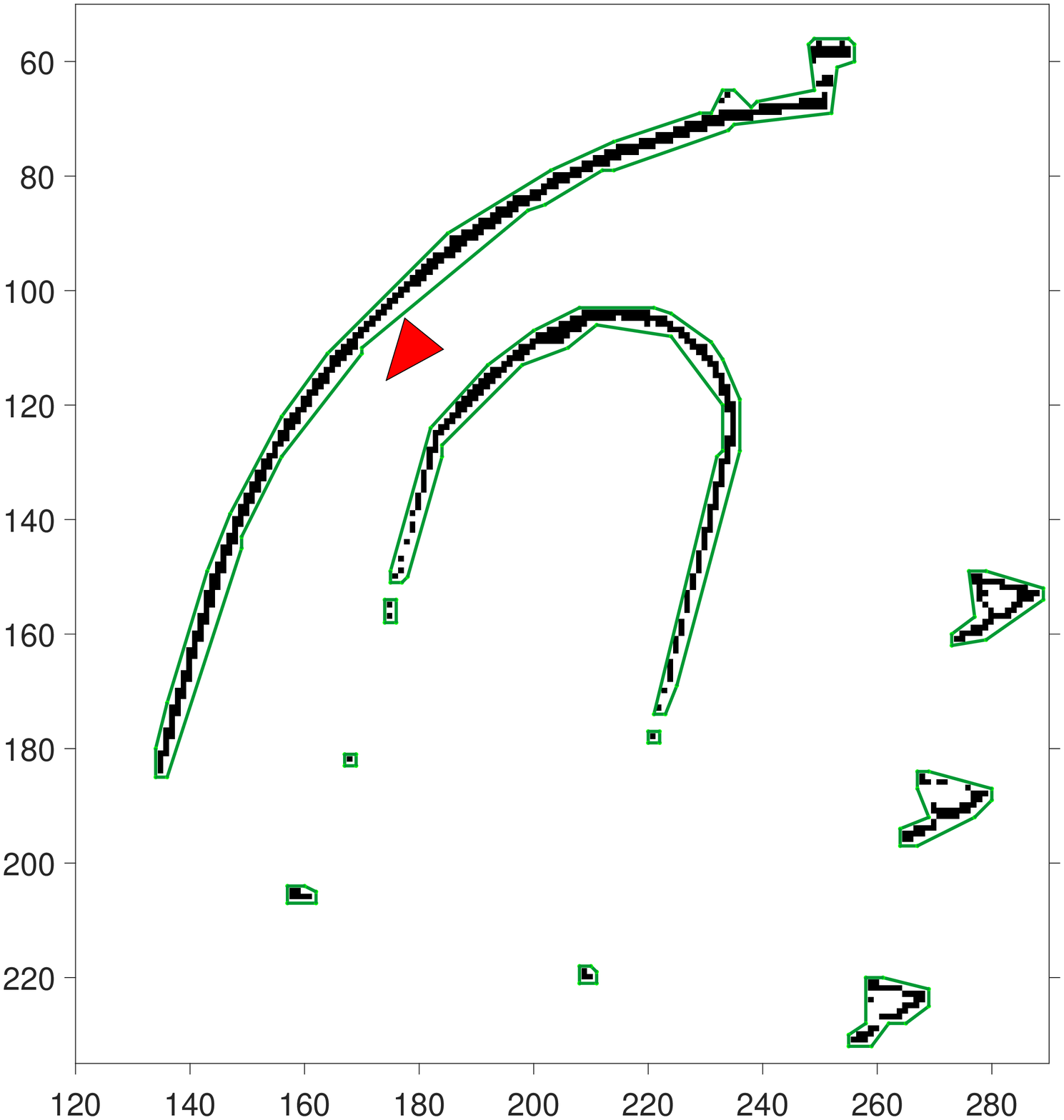}}
	\centering
	\subfigure[Final convex polygons]{\includegraphics[width=5.8cm]{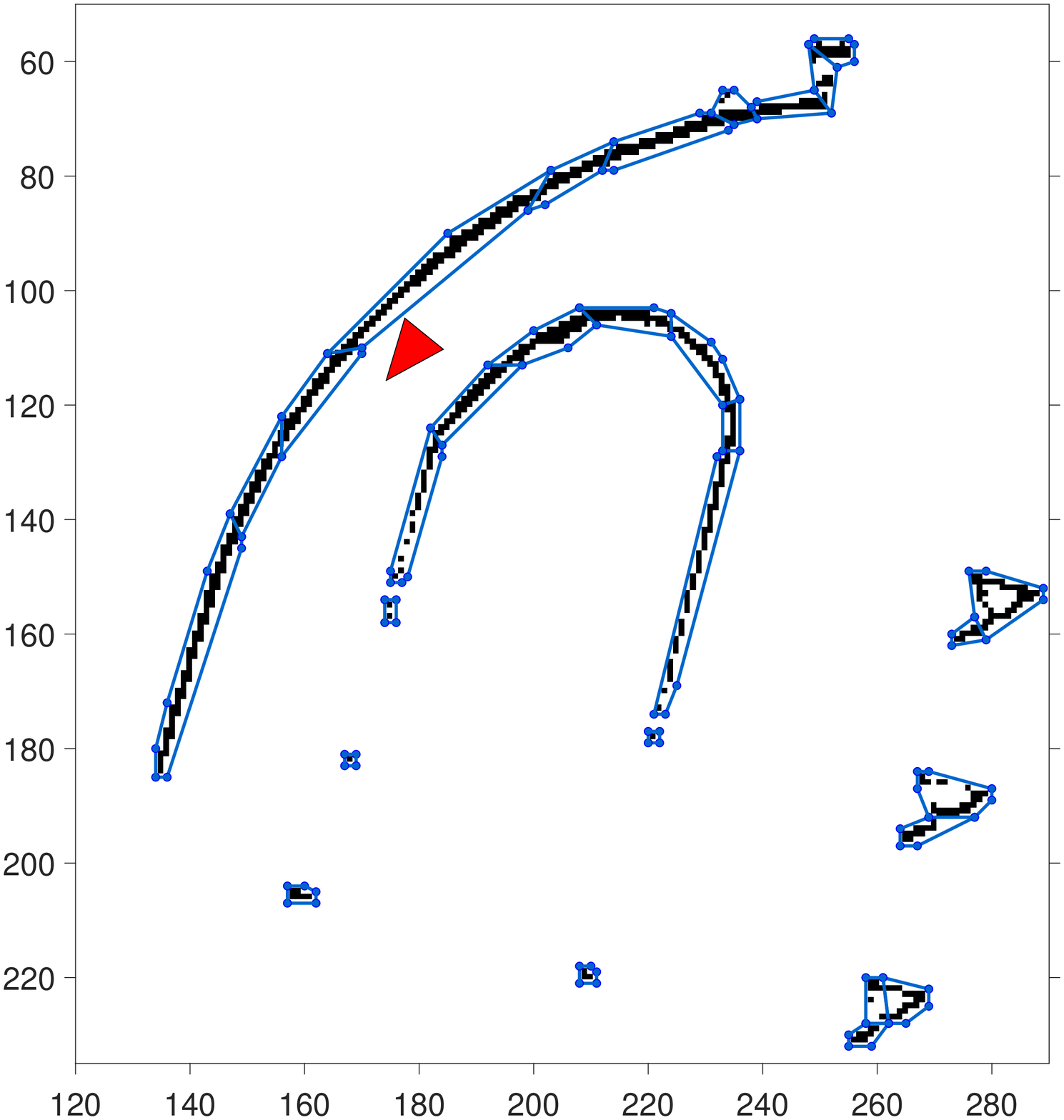}}
	\caption{The transformation process from the occupancy grid map to final convex polygons. The black pixels in (a) represent the static obstacle information, and the corresponding boundary information are denoted by the red polylines. (b) and (c) show the results after the process double threshold-based boundary simplification and convex polygon segmentation, respectively.}
	\label{process_two}
\end{figure*}

Accordingly, the experimental evaluation part can be divided into qualitative evaluation and quantitative evaluation. For the former, we describe the utility of the proposed approach, and show the obstacle detection ability of the proposed approach in several typical application scenarios; For the latter, we will quantitatively evaluate the performance of the proposed approach in terms of the simplification of obstacle representation. Specially, in practical applications, the representation method of obstacles is important for message transmission and motion planning. Compared with traditional occupancy grid map-based obstacle representation methods, the proposed approach makes use of the smaller amount of boundary points to represent the local static environments.

\begin{figure}[!htb]
	\centering
	\includegraphics[width=6.6cm]{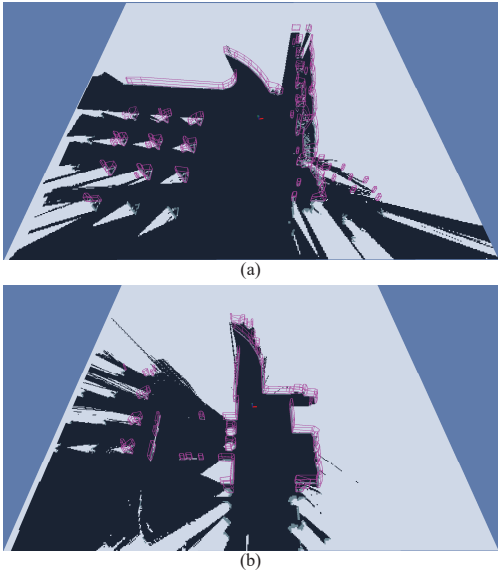}
	\caption{The vectorization representation results of local static environments. (a) In the uphill scene; (b) In the downhill scene.}
	\label{experiment_two}
\end{figure}

\subsection{The utility of vectorization representation}

At first, the transformation process of local environment representation is described, that is, the transformation process from the occupied grid map to final convex polygons, which also goes through boundary extraction and simplification. Without loss of generality, we take the local environment where the vehicle is going uphill (Fig. \ref{experimental_setup}(e)) as an example, the occupancy grid map and the final convex polygons are shown in Fig. \ref{process_one}. At the same time, Fig. \ref{process_two} describe the detailed process of convex polygon generation.

As shown in Fig. \ref{process_two}, compared with the traditional method of using occupancy grid map (Fig. \ref{process_two}(a)) to represent obstacle information, the cost of environment description can be greatly reduced by using convex polygons (Fig. \ref{process_two}(c)) to describe the local environment. At the same time, it overcomes the planning challenges brought by concave polygons (Fig. \ref{process_two}(b)).

On the above basis, the vectorization representation results of several typical scenes are shown in Fig. \ref{experiment_two}, and the purple convex polygons represent the local static environment information, while detailed static obstacle information is displayed through the occupancy grid map.

\begin{figure}[!htb]
	\centering
	\includegraphics[width=6.8cm]{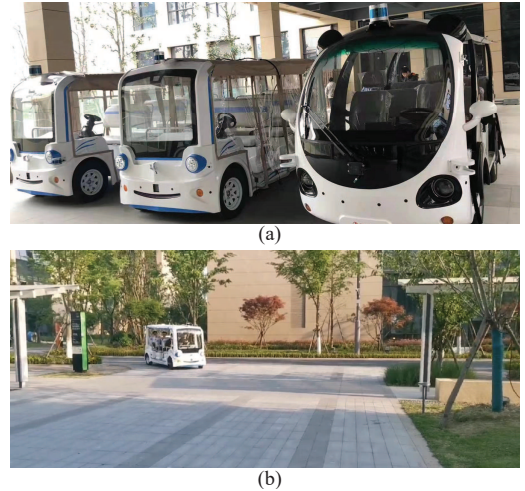}
	\caption{The driverless project in the park. (a) Unmanned ground vehicles applied in the project; (b) The vehicle during normal operation.}
	\label{project_setup}
\end{figure}

\subsection{Qualitative experimental results}
In fact, the proposed approach of local static environment perception is applied in our driverless project in the park, which is good at general obstacle detection, like the fallen traffic cone, the bizarre truck and other difficult object detection tasks for learning-based methods. In this section, the qualitative experimental results on typical scenes are given to verify the utility of the proposed approach in terms of the effectiveness and robustness of general obstacle detection. As shown in Fig. \ref{project_setup}, the proposed general obstacle detection approach has been applied in the project of park unmanned vehicles under normal operation.

\begin{figure*}[!htb]
	\centering
	\includegraphics[width=16cm]{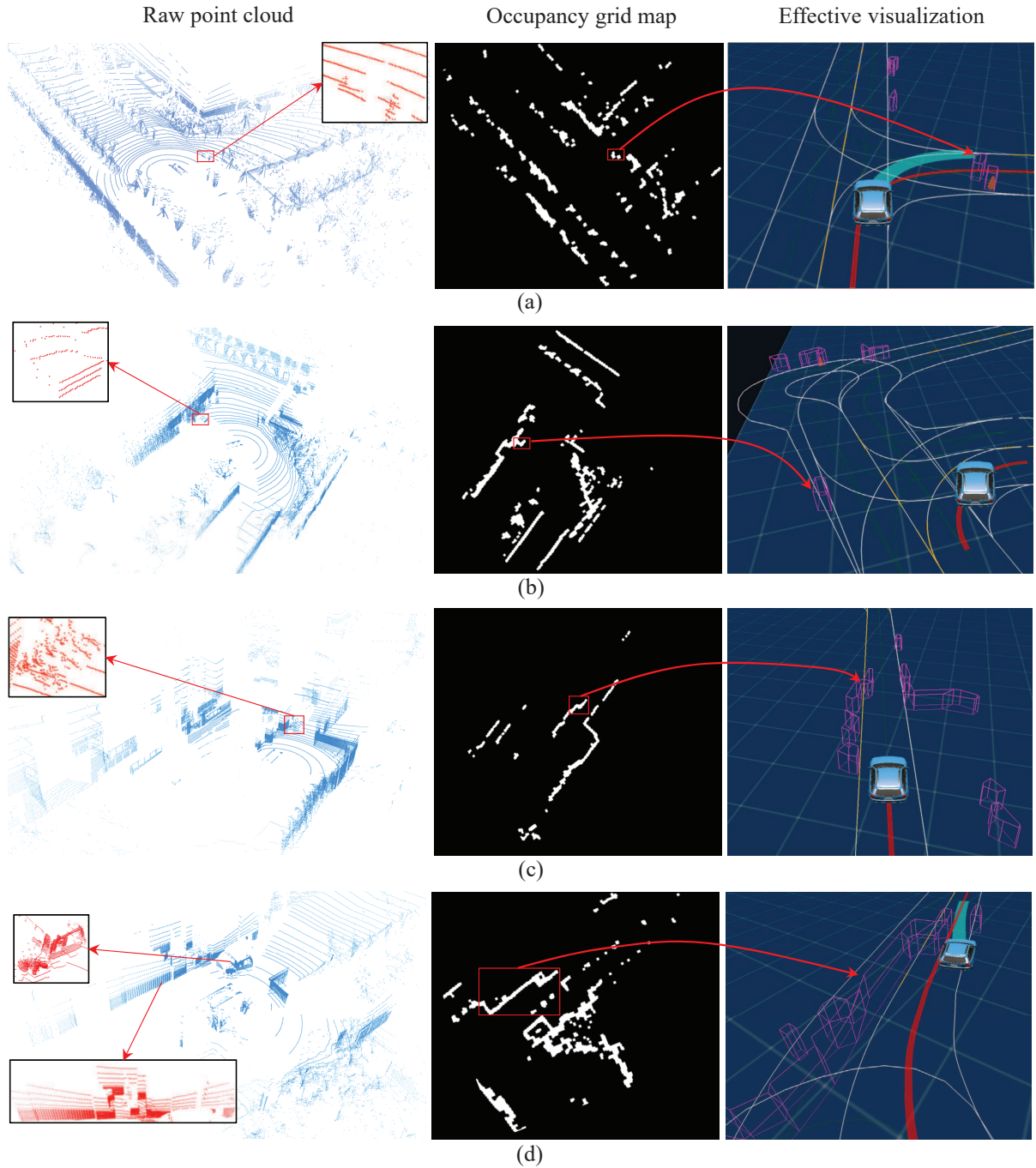}
	\caption{Four typical cases during the daily operation of unmanned ground vehicles.}
	\label{experiment_one}
\end{figure*}

During the daily operation of unmanned ground vehicles in the park, several typical cases (Fig. \ref{experiment_one}) have been collected to show the effectiveness of the proposed general obstacle detection approach. Compared with the learning-based obstacle detection methods, the proposed approach could avoid introducing the white list of static obstacles and complete the detection of general obstacles, even if the detected obstacles has never been seen before. Accordingly, the following will briefly introduce the typical cases.

In Fig. \ref{experiment_one} (a), the left figure shows the point cloud around the unmanned vehicle. On the right side of the vehicle, two traffic cones were placed, and one fell down. In addition, the local content of the corresponding point cloud is magnified, which are marked by the black rectangle in Fig. \ref{experiment_one}(a). For the learning-based method, the normally placed traffic cone can be recognized (show as the traffic cone icon in the right figure in Fig. \ref{experiment_one}(a)), while the fallen traffic cone cannot be detected. However, for the proposed approach, both two traffic cones could be detected, which are marked by the convex polygons in the right figure.

Then, the detection of the general obstacle for the waste box in the roadside is shown in Fig. \ref{experiment_one}(b). And the detailed point cloud and the occupancy grid map are shown in the left figure and the center figure, respectively. After the process of vectorization representation, the waste box could be represented as the convex hull in the right figure.

For the learning-based method, the single bicycle could be easily detected. Unfortunately, when multiple bicycles gathered together, it is difficult to detect every bicycle. Under the above circumstances, it is dangerous for the vehicle, because it is very likely to miss the detection of some bicycles and cause a vehicle accident. For our proposed approach, it is not necessary to detect every bicycle, but only to detect some areas are not accessible for the vehicle, which are marked by the multiple convex polygons, and the detailed information is shown in Fig. \ref{experiment_one}(c).

While compared with the detection of the car, the detection of the truck is more challenging, since the shape and size of different trucks vary greatly. Fig. \ref{experiment_one}(d) show the detection results of the proposed approach, and the detailed point cloud of large and small trucks are magnified in the left figure. In the center figure, the corresponding area occupied by trucks are marked by the red rectangle, and the final detection results are shown in the right figure.

Therefore, through the above four typical cases, the effectiveness of the proposed approach of general static obstacle detection is verified. In particularly, when there are strange static obstacles that have not been seen before around the autonomous vehicle, which are difficult for learning-based methods to be recognized and detected. However, the proposed approach could solve the above difficulties.

\subsection{Comparative quantitative evaluation}

\begin{figure}[!htb]
	\centering
	\includegraphics[width=8.5cm]{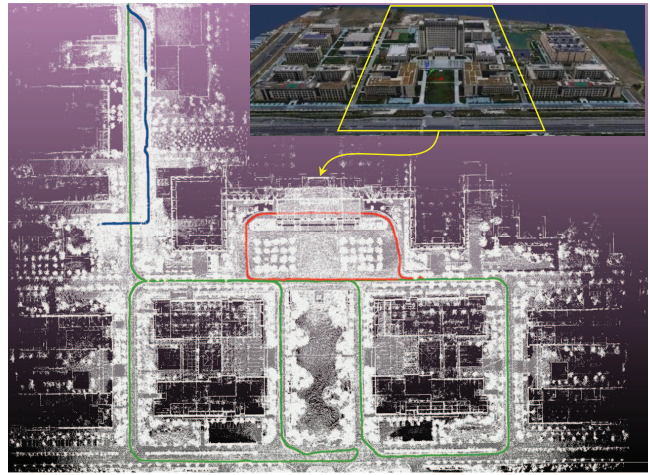}
	\caption{Experimental environment. The structural information is described by the white point cloud, and the red, green and blue lines represent threes trajectories of the vehicle.}
	\label{experiment}
\end{figure}

\begin{figure*}[!htb]
	\centering
	\subfigure[Dataset-1]{\includegraphics[width=5.9cm]{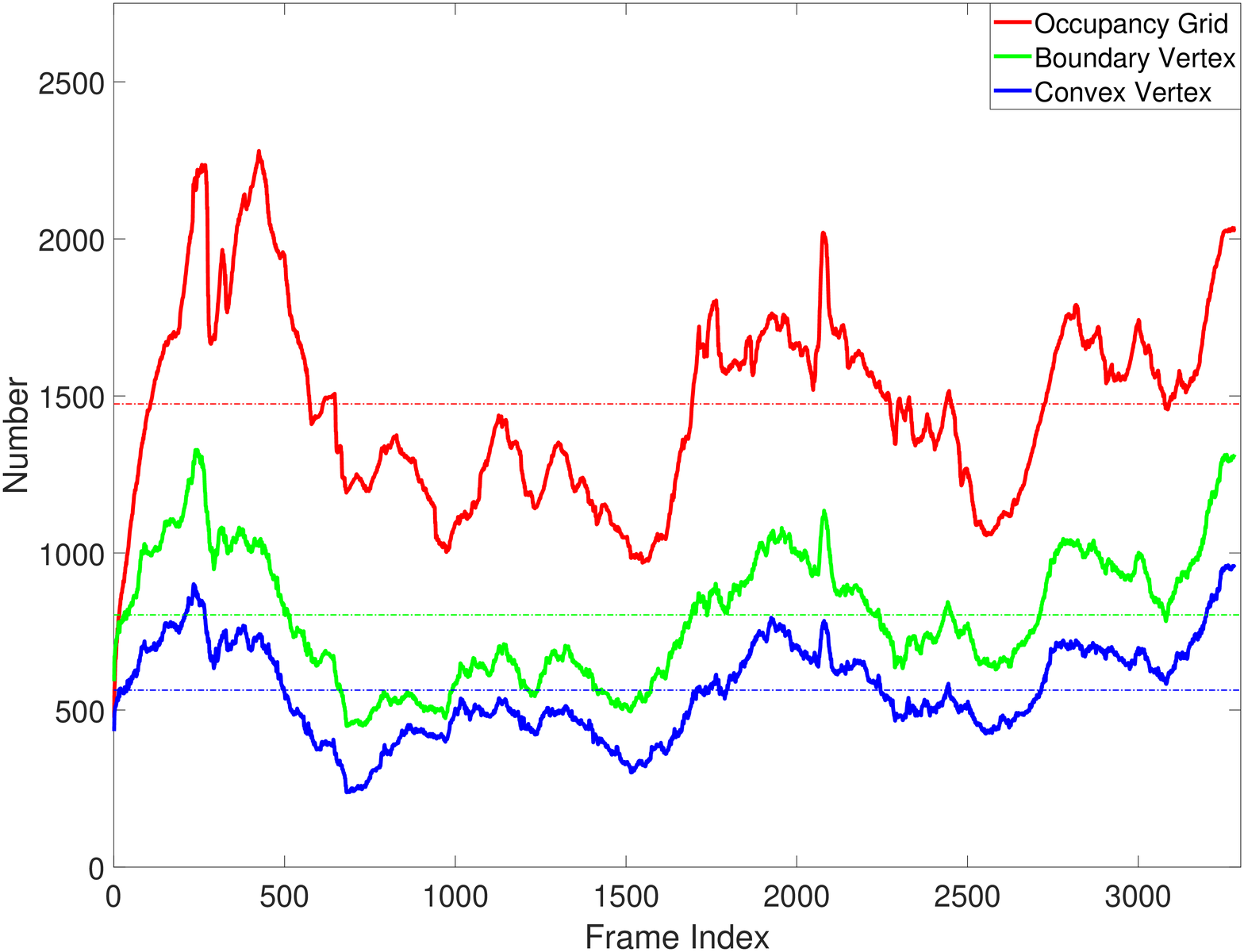}}
	\centering
	\subfigure[Dataset-2]{\includegraphics[width=5.9cm]{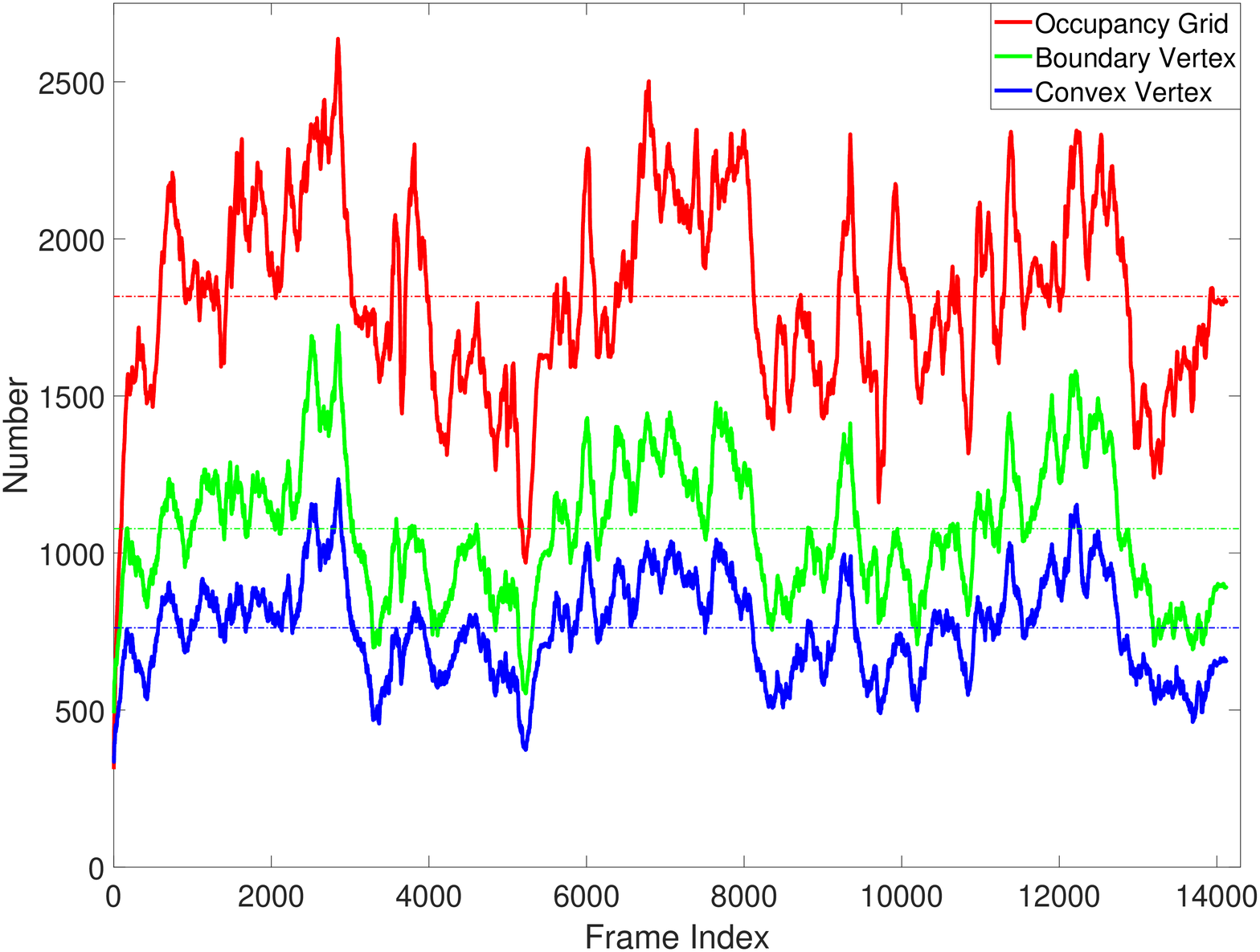}}
	\centering
	\subfigure[Dataset-3]{\includegraphics[width=5.9cm]{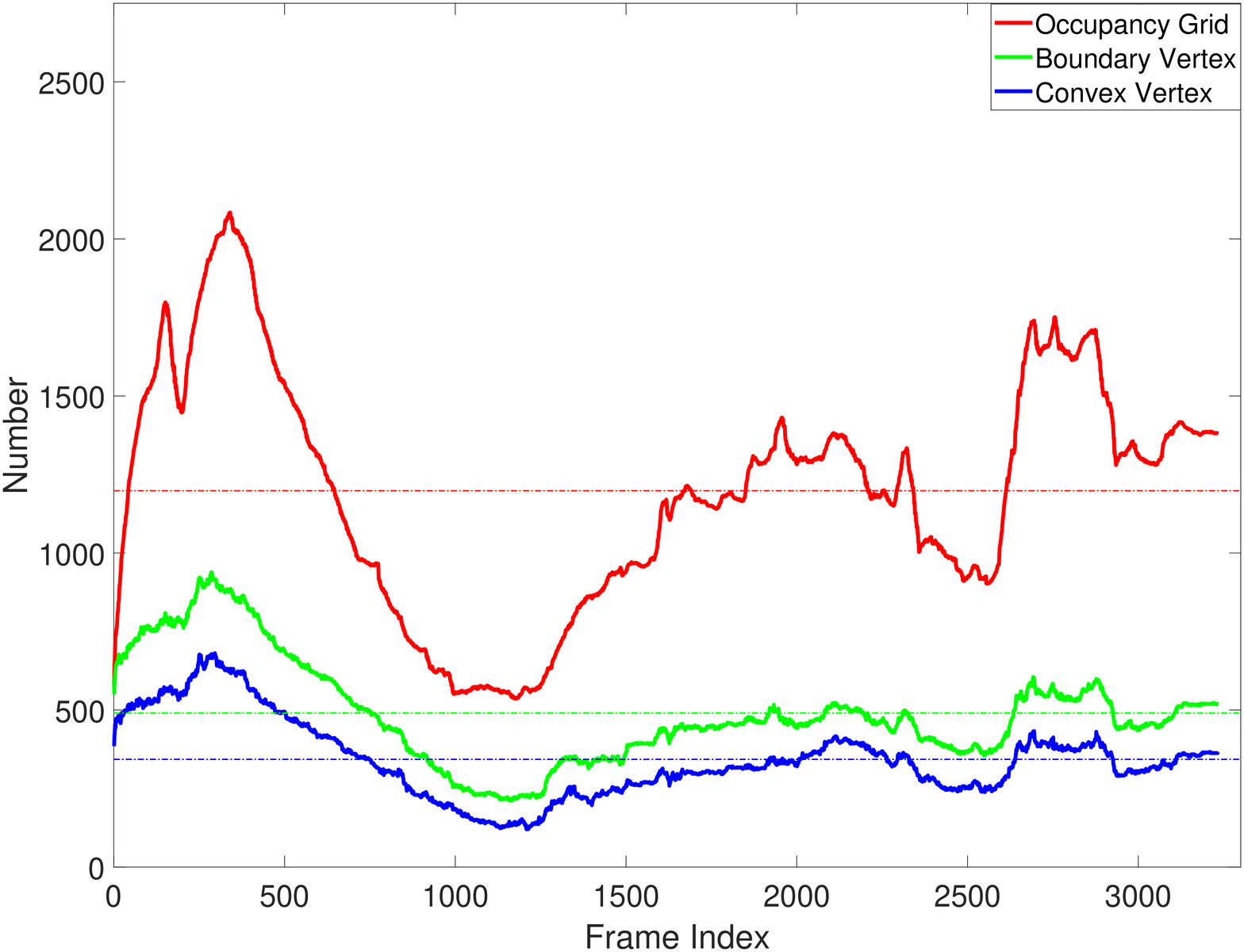}}
	\caption{The number of occupancy grids, boundary vertices and polygon vertices as the vehicle moves on. In addition, the average number is dotted by the dotted lines.}
	\label{num_one}
\end{figure*}

In addition, in order to show superior performance of the proposed approach, we have also carried out comparative analysis experiments through the vehicle platform (Fig. \ref{experimental_setup}(a)), and the corresponding environment and point cloud map are shown in Fig. \ref{experimental_setup}(b)-(e) and Fig. \ref{experiment}, respectively. In Fig. \ref{experiment}, the red, green and blue lines represent threes trajectories of the vehicle, respectively, which are used for dataset collection (\textcolor[rgb]{0.00,0.00,0.00}{Dataset-1}, \textcolor[rgb]{0.00,0.00,0.00}{Dataset-2}, \textcolor[rgb]{0.00,0.00,0.00}{Dataset-3}).

Through making use of multiple convex polygons to represent the local static environment, fewer points are required and less information is achieved for environmental representation compared with traditional methods based on the grid map, which is meaningful and improve the efficiency of information transmission and motion planning. Therefore the comparative analysis experiments are carried out, specially, for each test dataset, the number of occupancy grids in the local grid map, the number of vertices after boundary extraction and the number of polygon vertices after vectorization representation are counted, respectively. Note that the effective characterization information of the grid map are represented by multiple occupancy grids, since the number of occupancy grids are fewer than the free/unknown grids. The experimental results are shown in Fig. \ref{num_one}, and the three subfigures from left to right correspond to three datasets in turn, while the red, green and blue curves represent the the number of occupancy grids, boundary vertices and polygon vertices as the vehicle moves on, respectively.

\begin{figure}[!htb]
	\centering
	\includegraphics[width=7.5cm]{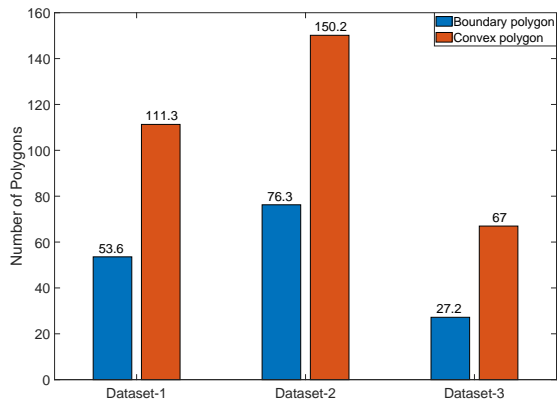}
	\caption{The average number of boundary and convex polygons for each dataset.}
	\label{num_two}
\end{figure}

Through the above comparative experimental results in terms of the number of grids or vertices, it is seen that the proposed approach does not need to retain every occupancy grid in the local map, and the corresponding number decreases $60.80\%$, $58.06\%$ and $71.37\%$, respectively. At the same time, compared with the number of vertices obtained by boundary extraction, although more polygons (Fig. \ref{num_two}) are used to represent the local environment (single concave polygon will be segmented into multiple convex polygons), there are fewer polygon vertices for the proposed approach. This is mainly due to the effectiveness of the applied double-threshold boundary simplification.

\begin{figure}[!htb]
	\centering
	\includegraphics[width=8cm]{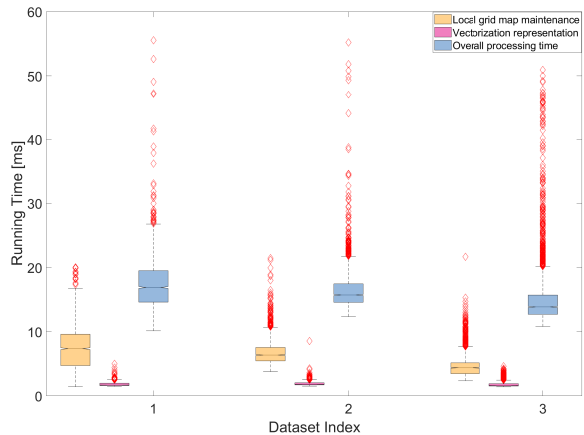}
	\caption{The running time performance of single frame for the proposed approach.}
	\label{time}
\end{figure}

On the other hand, running time as an important performance term is also analyzed. We have collected the running time information of the first 3000 frames for each dataset, and the running time performance is reflected in three aspects: local grid map maintenance, vectorization representation and overall processing time. The experimental statistical results are shown in Fig. \ref{time}, except for a few outliers marked by the red diamond, the overall processing time of single frame is concentrated in $15ms$, while the running times for the local grid map maintenance and the vectorization representation are stable within $10ms$ and $5ms$, respectively. Therefore the performance of running time shows that the proposed approach can be used for real-time local static environment perception.

\section{Conclusion and future work}
In this work, a novel and compact vectorization representation approach of local static environments has been proposed for unmanned ground vehicles equipped with the 3D LiDAR sensor. At first, the vehicle pose and the obstacle point cloud are obtained via pose estimation and ground segmentation, respectively. Then, grid map generation is carried to obtain the final grid map information for local static environments. Finally, several convex polygons are used to represent the local static environment through the vectorization representation.

Our proposed approach is currently being applied in our driverless project in the park, and the typical scenarios verify the effectiveness and robustness of general obstacle detection. Furthermore, the quantitative evaluation shows the superior performance on the effective representation for the local static environments compared with the traditional grid map-based methods, which is based on less information. And the performance of running time shows that the proposed approach can be used for real-time local static environment perception.

In the future work, we will propose a unified obstacle detection framework for autonomous vehicles, which will combine with the deep learning-based approaches and the advantage of grid map on general obstacle detection.

\begin{IEEEbiography}[{\includegraphics[width=1in,height=1.25in,clip,keepaspectratio]{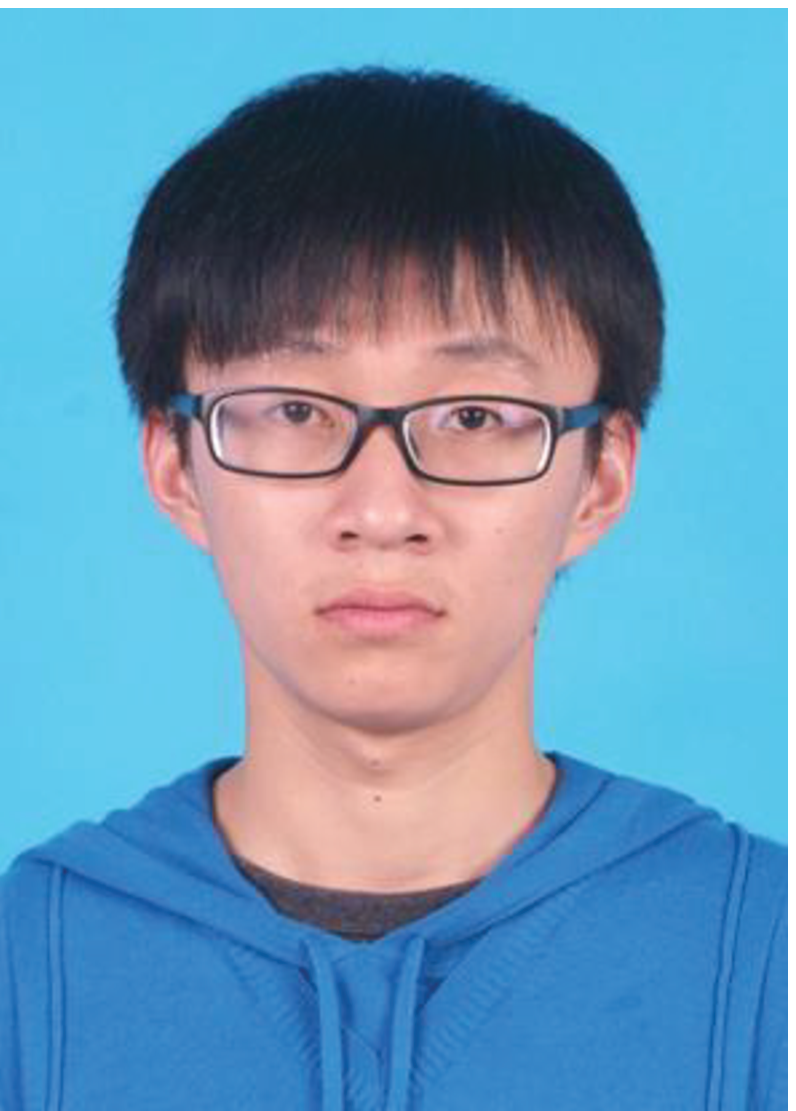}}]{Haiming Gao} received the B.S. degree in mechanical engineering and automation from Hohai University, Nanjing, China, in 2015, and the Ph.D. degree in control science and engineering from Nankai University, Tianjin, China, in 2020. From 2020 to 2021, he worked as a Senior Algorithm Engineer in IAS BU (Huawei Intelligent Automotive Solution), with research interests include freespace extraction, APA(Auto Parking Asist) and AVP(Automated Valet Parking). He is currently a Research Fellow with Zhejiang Lab.
	
	His research interests include localization and perception for intelligent vehicles and mobile robots.
\end{IEEEbiography}

\begin{IEEEbiography}[{\includegraphics[width=1in,height=1.25in,clip,keepaspectratio]{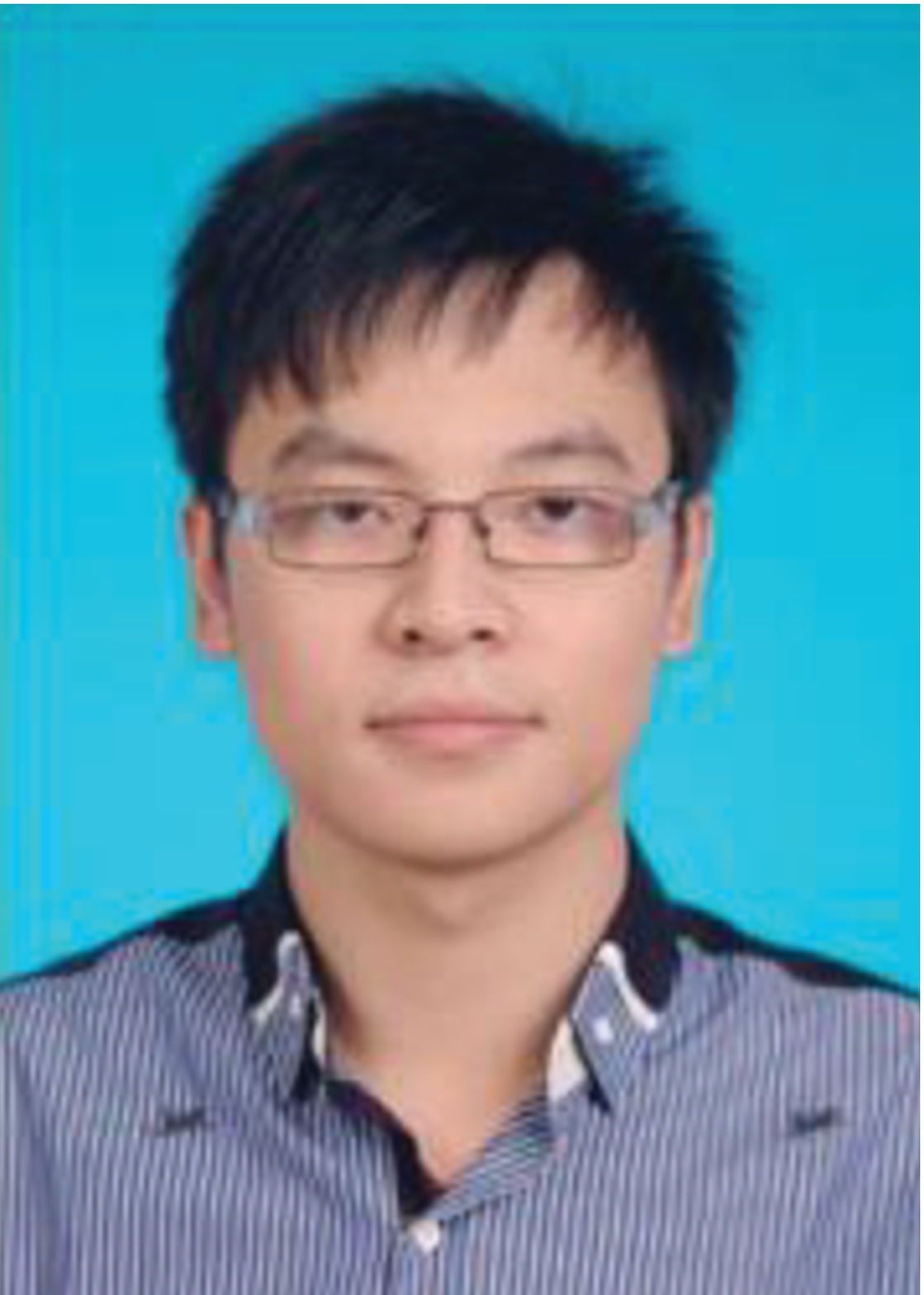}}]{Qibo Qiu} received the B.S. degree in software engineering at Xidian University, Xian, China, in 2014, and the M.S degree in computer science at Zhejiang University, Hangzhou, China, in 2017. From 2016 to 2018, he worked as  a senior engineer in DiDi research and Fabu tech with research interests include HDmap(high resolution map), localization, and perception.
	
	He is currently an engineer with Zhejiang Lab, his research interests include road marker detection, 2D/3D lane detection and perception for intelligent vehicles and mobile robots.
\end{IEEEbiography}

\begin{IEEEbiography}[{\includegraphics[width=1in,height=1.25in,clip,keepaspectratio]{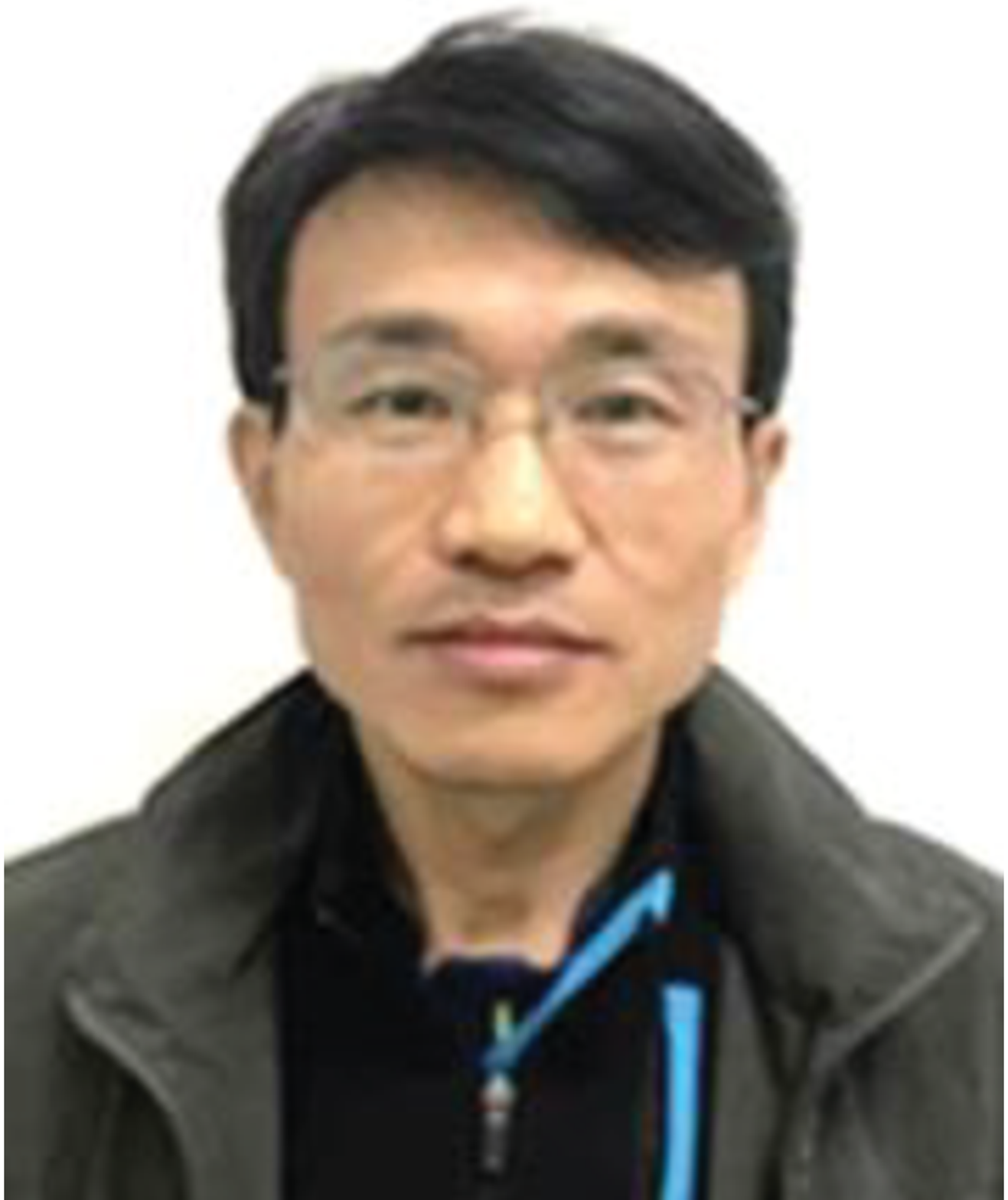}}]{Wei Hua} received the PhD degree in applied mathematics from Zhejiang University. He is currently a senior research expert of Zhejiang Laboratory.
	
	His current research interests include autonomous driving, intelligent simulation, digital twin, reinforcement learning and AI-based algorithms, and related works have been applied to intelligent transportation, smart city, smart community, etc.
	
\end{IEEEbiography}

\begin{IEEEbiography}[{\includegraphics[width=1in,height=1.25in,clip,keepaspectratio]{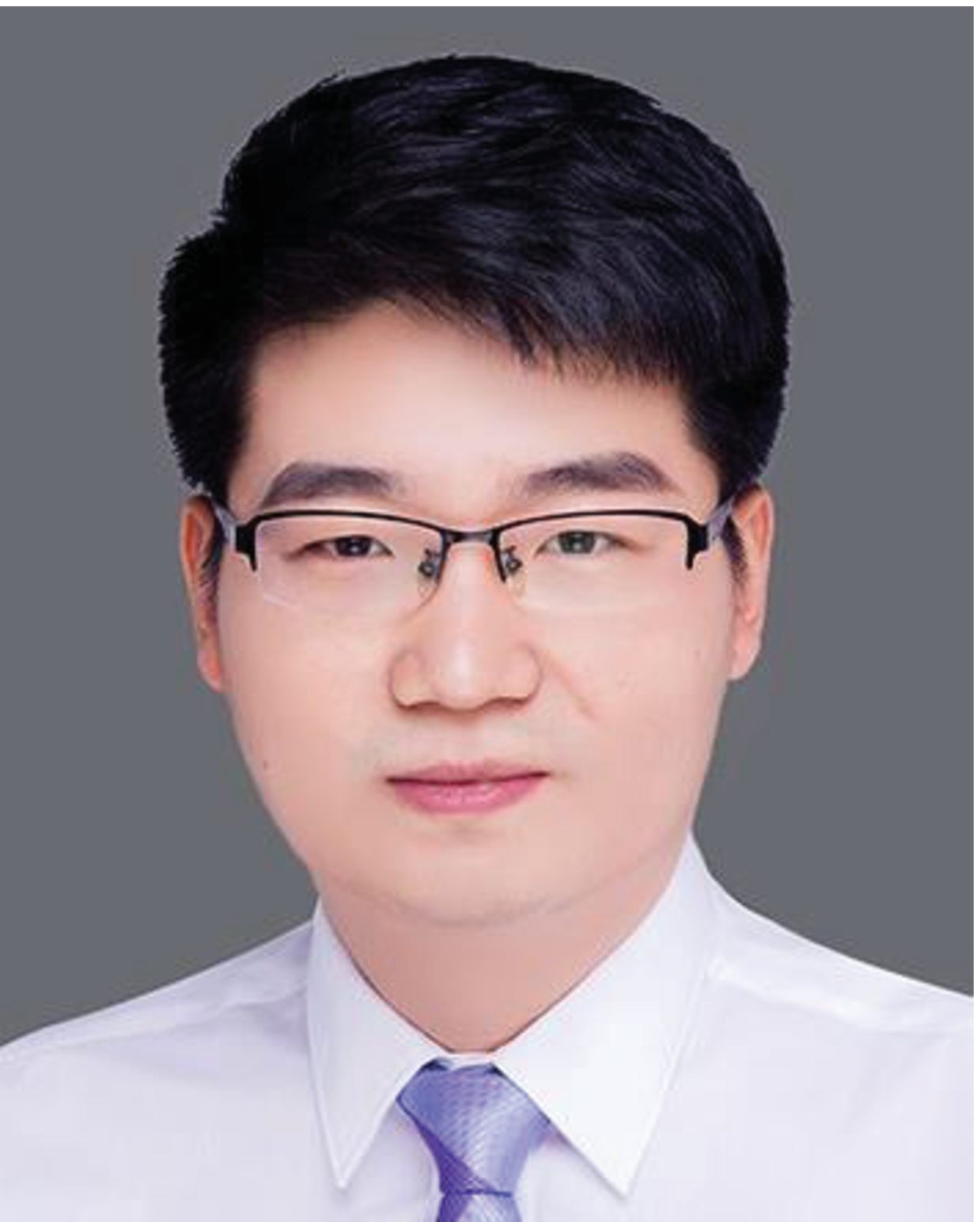}}]{Xuebo Zhang} (M'12$-$SM'17) received the B.Eng. degree in automation from Tianjin University, Tianjin, China, in 2006, and the Ph.D. degree in control theory and control engineering from Nankai University, Tianjin, China, in 2011.
	
	From 2014 to 2015, he was a Visiting Scholar with the Department of Electrical and Computer Engineering, University of Windsor, Windsor, ON, Canada. He was a Visiting Scholar with the Department of Mechanical and Biomedical Engineering, City University of Hong Kong, Hong Kong, in 2017. He is currently a Professor with the Institute of Robotics and Automatic Information System, Nankai University, and Tianjin Key Laboratory of Intelligent Robotics, Nankai University. His research interests include planning and control of autonomous robotics and mechatronic system with focus on time-optimal planning and visual servo control; intelligent perception including robot vision, visual sensor networks, SLAM; reinforcement learning and game theory.
	
	Dr. Zhang is a Technical Editor of \emph{IEEE/ASME Transactions on Mechatronics} and an Associate Editor of \emph{ASME Journal of Dynamic Systems, Measurement, and Control}.
\end{IEEEbiography}

\begin{IEEEbiography}[{\includegraphics[width=1in,height=1.25in,clip,keepaspectratio]{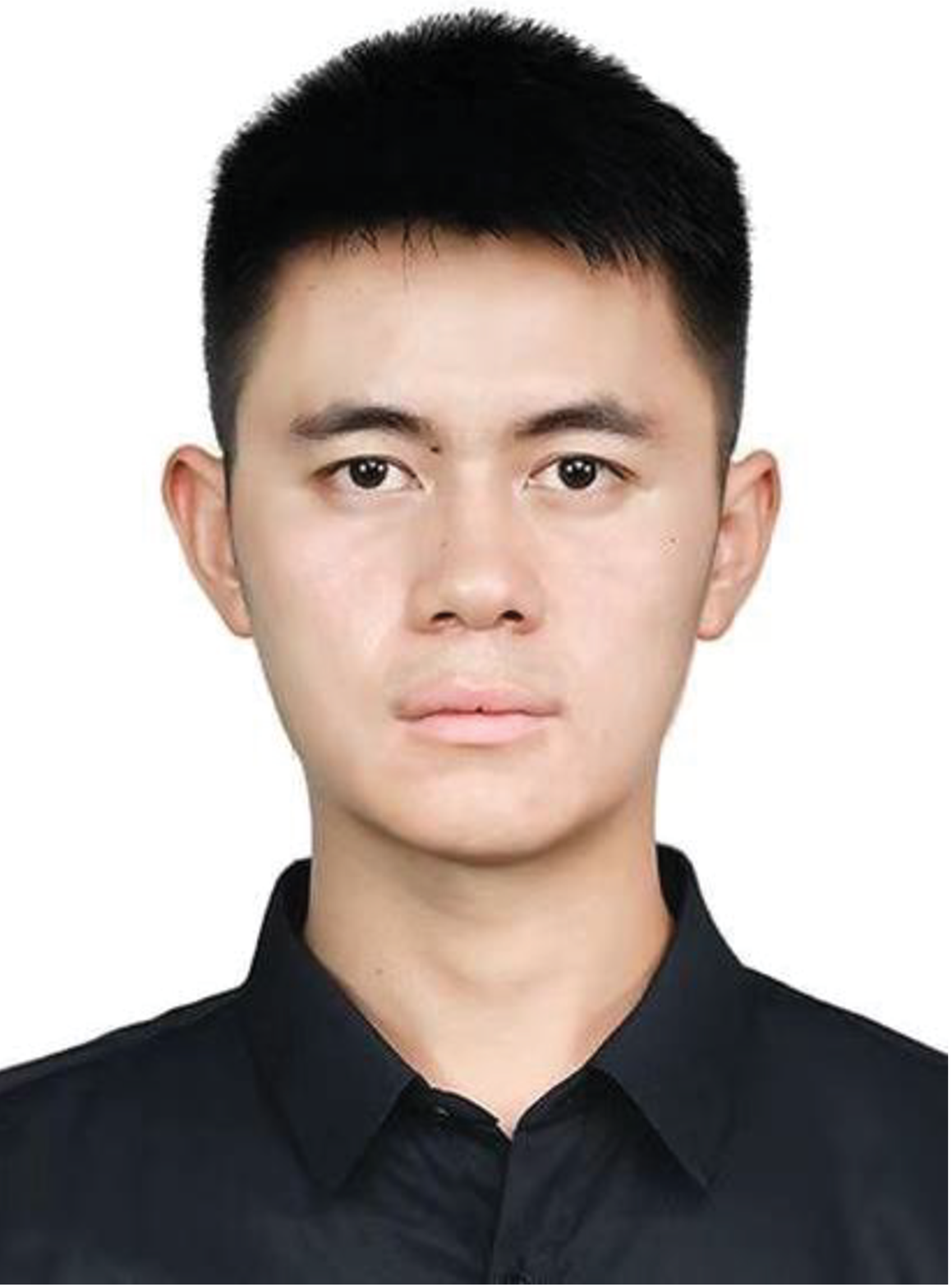}}]{Zhengyong Han} received the B.S. degree in automobile engineering from Hunan University, Changsha, China, in 2014, and M.D. degree in mechanical and electronic engineering from Shenyang Institute of Automation, Chinese Academy of Sciences, Shenyang, China, in 2017. From 2017 to 2019, he works as System Architecture Engineer in ATDSAIC MOTOR, with work focus on system scheme and integration on L4 intelligent truck. He is currently a Senior Engineer Commissioner in Zhejiang Lab.
	
	His research interests include system architecture and integration for intelligent vehicle.
\end{IEEEbiography}

\begin{IEEEbiography}[{\includegraphics[width=1in,height=1.25in,clip,keepaspectratio]{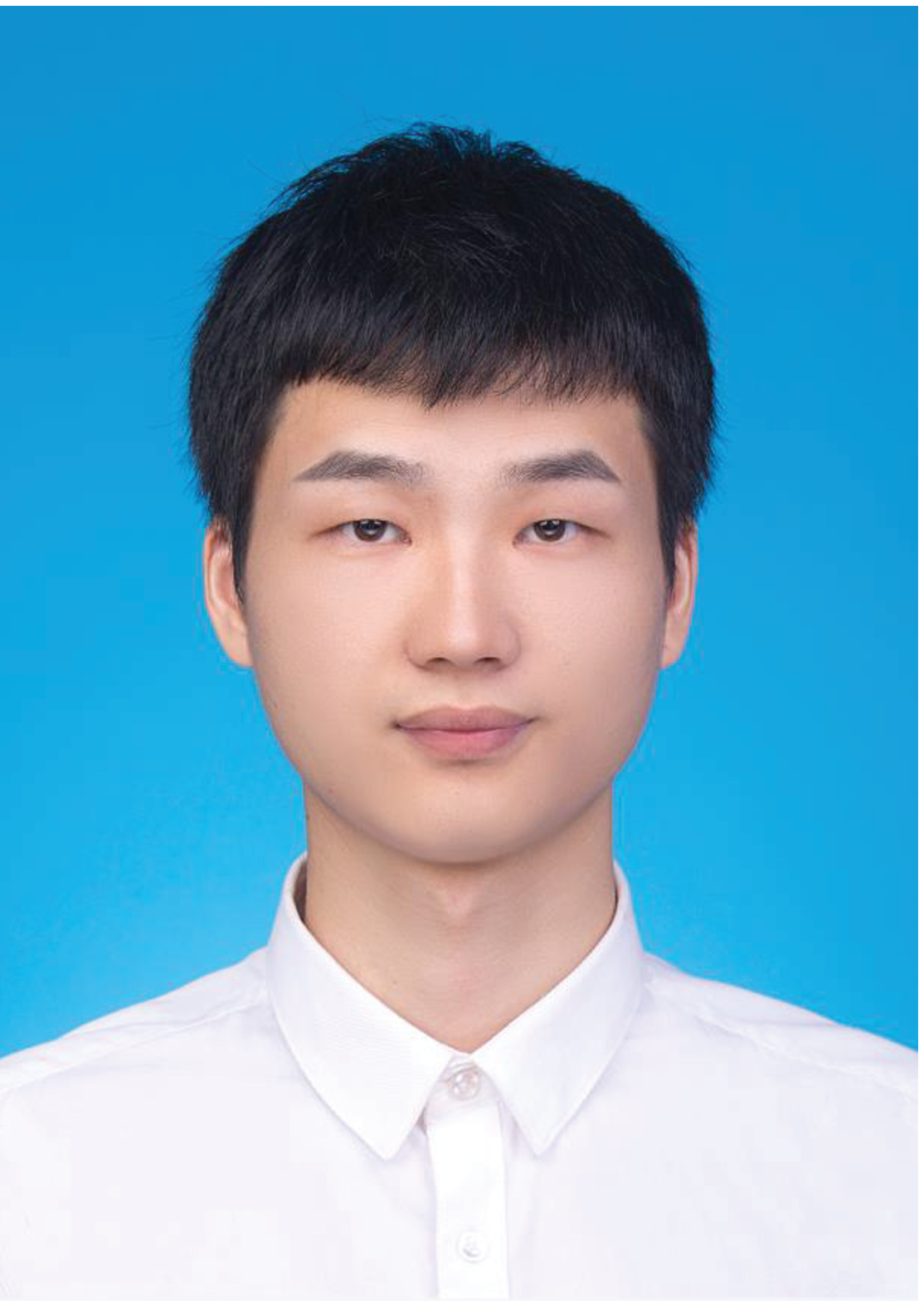}}]{Shun Zhang} received the B.Eng. and the M.Eng. degrees from Wuhan University, Wuhan, China in 2017 and 2019, respectively. He is currently a Algorithm Engineer with Zhejiang Lab. His research interests include SLAM(Simultaneous Localization and Mapping), GPR(General Place Recognition) and perception for intelligent vehicles and mobile robots.
\end{IEEEbiography}

\end{document}